%% file: acl_latex.tex
\documentclass[11pt]{article}

\usepackage[preprint]{acl}

\usepackage{times}
\usepackage{latexsym}

\usepackage[T1]{fontenc}

\usepackage[utf8]{inputenc}

\usepackage{microtype}

\usepackage{inconsolata}

\usepackage{graphicx}

\usepackage{graphicx} 
\usepackage{algorithm}
\usepackage{algpseudocode}
\usepackage{multirow}
\usepackage{booktabs}
\usepackage{arydshln}
\usepackage{xcolor,colortbl}
\usepackage{color}
\usepackage[most]{tcolorbox}
\usepackage{amsmath}
\usepackage{amssymb}
\usepackage{amsfonts}
\usepackage{subcaption}
\usepackage{float}
\usepackage{wrapfig} 
\usepackage{enumitem}
\usepackage{stfloats}

\newcommand{\think}[1]{\textcolor{blue}{\texttt{<think>}} #1 \textcolor{blue}{\texttt{</think>}}}
\newcommand{\formula}[1]{\textcolor{cyan}{\texttt{<formula>}} #1 \textcolor{cyan}{\texttt{</formula>}}}

\newcommand{\answer}[1]{\textcolor{purple}{\texttt{<answer>}} #1 \textcolor{purple}{\texttt{</answer>}}}

\title{\textsc{MedCalc-R1}: Knowledge-Guided Reward Framework for Medical Mathematical Reasoning}

\author{Haotian Wang\thanks{Equal contribution}, Lian Yan$^*$, Xingzhi Yao, Fanshu Meng, Ye He, \\ 
\textbf{Jingchi Jiang, Yi Guan\thanks{Corresponding Author}}\\
  Harbin Institute of Technology \\
  \texttt{wanght1998@gmail.com}, \texttt{\{23b903008,22s103150,24s003058\}@stu.hit.edu.cn}, \\
 \texttt{yehe@ir.hit.edu.cn}, \texttt{\{jiangjingchi, guanyi\}@hit.edu.cn}
}

\begin{document}
\maketitle

\input{sections/abstract}
\input{sections/1_introduction}

\input{sections/2_related_work}

\input{sections/3_method}

\input{sections/4_experiment}

\input{sections/5_conclusion}
\input{sections/limitations}
\input{sections/ethics_statement}
\bibliography{custom}

\appendix
\input{sections/appendix}


\end{document}

%% file: sections/abstract.tex

\begin{abstract}
    In Reinforcement Learning with Verifiable Rewards (RLVR) frameworks for mathematical reasoning tasks, floating-point results are typically evaluated using a tolerance-based reward. However, this strategy suffers from challenges such as difficulty in threshold calibration, unstable training dynamics, and limited accuracy, especially in clinical scenarios. To address these limitations, we propose a knowledge-guided hybrid reward framework (\textsc{MedCalc-R1}). Specifically, we introduce a knowledge verification reward mechanism that enforces explicit generation of computational formulas, which are further validated by an external verifier to enhance interpretability and reasoning reliability. Furthermore, we design a hybrid soft-hard reward scheme combining a hard constraint based on clinical safety thresholds with a soft, precision-sensitive reward that progressively guides learning within the acceptable range. Experimental results demonstrate that our method significantly outperforms existing baselines in both reasoning accuracy and generalization capability, validating the effectiveness and applicability in safety-critical domains. \textit{Code will be released upon acceptance.}
\end{abstract}

%% file: sections/1_introduction.tex
\section{Introduction}
In recent years, Large Language Models (LLMs) have demonstrated remarkable capabilities in complex reasoning tasks such as mathematical problem solving~\citep{shao2024deepseekmath, yang2024qwen2}, logical inference~\citep{yang2024harnessing, qin2024relevant, liu2025logical}, and program synthesis~\citep{austin2021program, guo2024deepseek}. To further enhance the reliability of these models, Reinforcement Learning with Verifiable Rewards (RLVR)~\citep{guo2025deepseek,wang2024reinforcement} has emerged as a promising paradigm. By integrating symbolic verifiability with rule-based supervision, RLVR leverages objective binary feedback from logical rules or execution environments to guide models in mastering multi-step reasoning capabilities. This approach bridges the gap between neural generation and symbolic correctness, laying a critical foundation for deploying LLMs in high-stakes and precision-demanding scenarios~\citep{xu2025towards,he2025breaking,wang2025survey}.

\begin{figure}[]
\centering
\includegraphics[width=\linewidth]{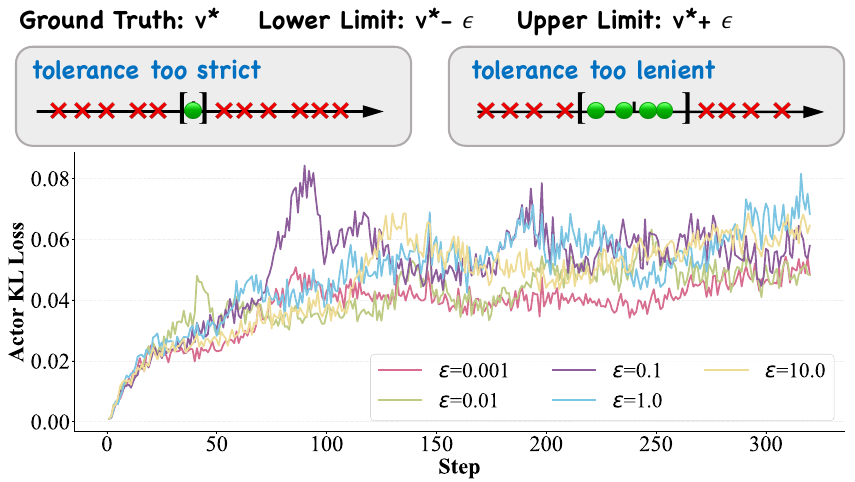}
\vspace{-6mm}
\caption{Sensitivity analysis of tolerance boundaries. A strict tolerance explicitly inhibits convergence speed, keeping the curve low. In contrast, excessive relaxation causes high-variance oscillations, destabilizing the gradient optimization process.}
\label{fig:introduction}
\end{figure}


However, applying RLVR to tasks involving floating-point outputs presents unique challenges compared to discrete symbolic domains. Existing methods~\cite{tan2025reasonrft,liu-etal-2025-compassverifier} predominantly adopt a tolerance-based binary reward, where a prediction is deemed correct solely if it falls within a predefined margin of the ground truth. Nevertheless, this heuristic introduces a fundamental trade-off between stability and precision.. First, \textbf{\textit{the sensitivity to tolerance thresholds creates an optimization dilemma}}. As illustrated in Figure \ref{fig:introduction}, overly strict thresholds result in severe reward sparsity, causing policy stagnation where the model fails to capture sufficient feedback for effective updates. Conversely, excessively lenient thresholds introduce reward noise, triggering large gradient fluctuations and forcing the policy to deviate significantly from the reference distribution, thereby destabilizing the training process. Second, \textbf{\textit{this mechanism provides only weak, outcome-oriented supervision}}. It neglects the fidelity of the reasoning trajectory, frequently rewarding fortuitous approximations rather than logically sound derivations.

These limitations become prohibitive barriers in safety-critical clinical scenarios, where numerical precision is not merely a performance metric but a fundamental safety constraint~\citep{singhal2023large,thirunavukarasu2023large,lucas2024reasoning}. In tasks such as creatinine clearance-based drug dosing and individualized risk scoring, even minor deviations permitted by tolerance-based rewards can precipitate deleterious downstream effects. A slight miscalculation in these contexts can lead to incorrect dosage regimens, misdiagnoses, or life-threatening adverse events~\citep{cicero2020medication,hijji2025indispensable}. Although general LLMs have achieved high scores on standard mathematical benchmarks~\citep{jaech2024openai}, they fundamentally lack the rigor required for healthcare. Existing methods, with their reliance on opaque approximation, fail to satisfy the stringent requirements for accuracy, transparency, and reliability essential for real-world clinical deployment.

To address these challenges, we introduce a knowledge-guided hybrid reward framework (\textsc{MedCalc-R1}), which enhances reasoning fidelity and safety through two complementary mechanisms. First, to address the opacity of neural reasoning, we design a knowledge verification reward. This mechanism enforces the model to explicit generate formulas, which are then verified by an external solver for both structural correctness and intermediate consistency. Second, to tackle the trade-off between numerical precision and training stability, we propose a hybrid soft–hard reward scheme. This module integrates clinically valid tolerance intervals as hard constraints (ensuring safety) with progressive accuracy-based incentives as soft optimization signals (driving precision). By unifying symbolic process supervision with constraint-aware numerical optimization, our framework offers a robust paradigm for high-stakes medical reasoning.

The contributions of this work are summarized as follows:
\begin{itemize}[itemsep=2pt,topsep=2pt,parsep=2pt,leftmargin=*]
    \item We pioneer the integration of formula-level knowledge verification into the RLVR framework, which enforces explicit symbolic generation to enhance reasoning interpretability.
    \item We design a hybrid soft-hard reward scheme that integrates clinical safety constraints with progressive precision incentives, effectively balancing training stability with numerical accuracy.
    \item Experiment results demonstrate that our framework outperforms state-of-the-art baselines in both accuracy and robustness, showing superior generalization for clinical deployment.
\end{itemize}

%% file: sections/2_related_work.tex
\section{Related Work}


\paragraph{LLMs for Mathematical Reasoning.}
LLMs have demonstrated remarkable proficiency in complex reasoning tasks \citep{wang2025survey,zhang2025generative}. Advanced strategies, ranging from Chain-of-Thought (CoT) prompting \citep{wei2022chain,chu-etal-2024-navigate} to tool-augmented learning \citep{qu2025tool}, enable frontier models like GPT-4 and PaLM to perform effective multi-step deduction. To further enhance arithmetic precision, \citet{gao2023pal} employs Python code generation to solve complex problems, and \citet{imani2023mathprompter} explores diverse reasoning paths to improve robustness. Although these approaches achieve impressive performance on mathematical reasoning benchmarks~\citep{cobbe2021training,hendrycks2021measuring}, they predominantly rely on outcome-oriented verification, often neglecting the fidelity of the intermediate reasoning process. 


\paragraph{RL for LLMs Reasoning Tasks.} 
Reinforcement Learning (RL) has established itself as a pivotal paradigm for unlocking the reasoning potential of LLMs beyond Supervised Fine-Tuning (SFT) \citep{ouyang2022training}. To enhance logical coherence, prior works introduced process reward models, which provide dense step-level supervision~\citep{yuan2023rrhf,lightman2024lets}. However, the prohibitive cost of annotated reasoning traces limits their scalability~\cite{havrilla2024glore}. Consequently, recent research has pivoted towards RLVR, demonstrating that simple binary signals derived from rule-based verifiers are sufficient to elicit strong reasoning capabilities~\cite{guo2025deepseek,su2025crossing}. Nevertheless, unlike discrete symbolic domains where verification is deterministic, existing tolerance-based reward mechanisms for floating-point reasoning lack robustness, inevitably leading to optimization instability.

\begin{figure*}[]
\centering
\includegraphics[width=0.8\linewidth]{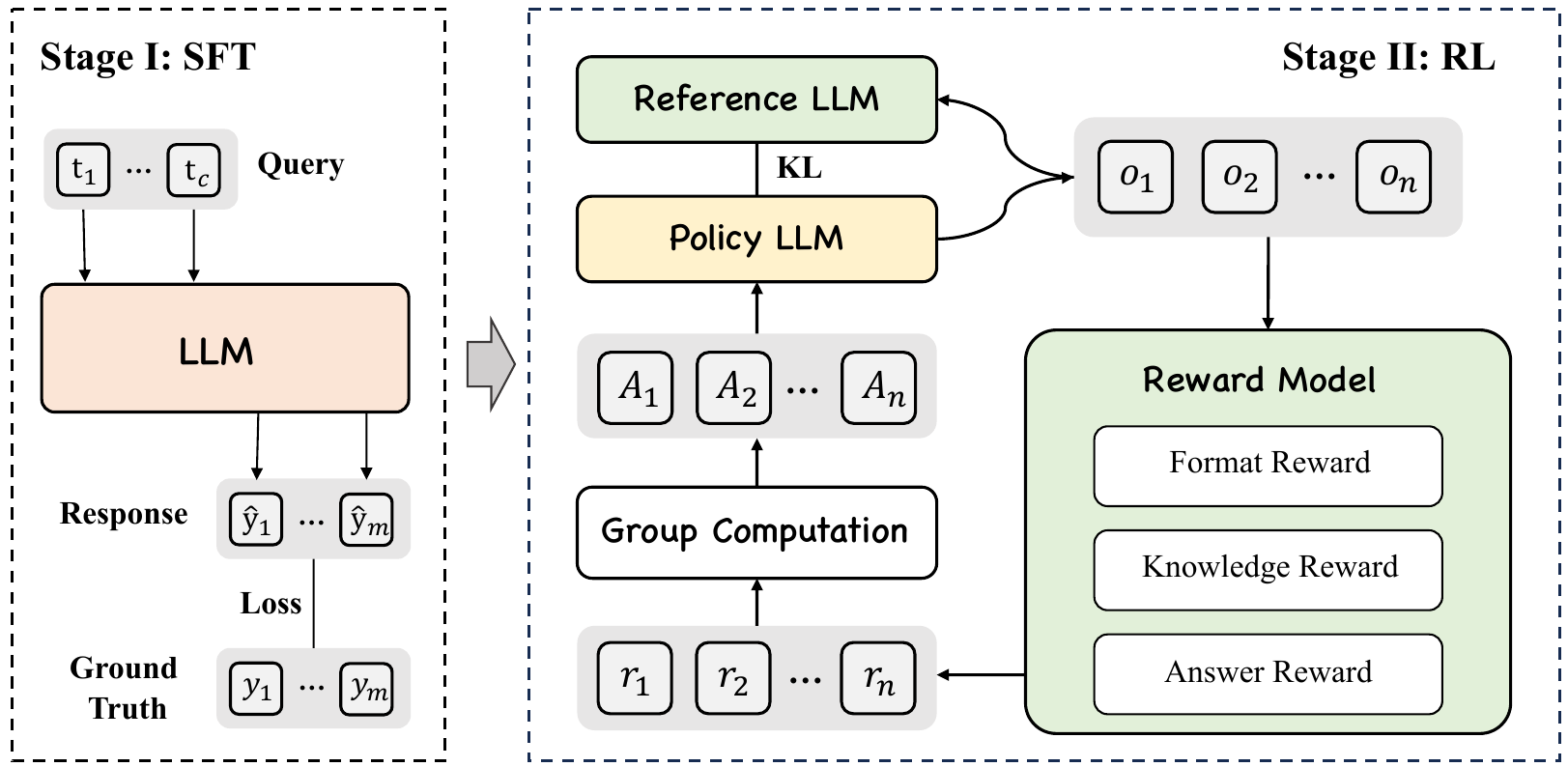}
\vspace{-2mm}
\caption{Overview of the proposed knowledge-guided reward framework. The framework follows a two-stage paradigm: Stage I (SFT): the model is trained to acquire output formatting and basic medical knowledge recall; Stage II (RL): candidate outputs are optimized using GRPO, where a reward model integrates format, formula, and answer rewards, while KL regularization ensures stable policy updates.
}
\label{fig:main}
\end{figure*}

\paragraph{LLMs for Medical Mathematical Reasoning.}  
Unlike general-purpose benchmarks \citep{cobbe2021training,hendrycks2021measuring}, clinical tasks function in high-stakes environments, where numerical errors can directly compromise patient safety~\cite{khandekar2024medcalcbench}. Consequently, clinical AI demands strict adherence to safety constraints and interpretability. While LLMs have revolutionized clinical NLP tasks, ranging from information extraction \citep{ruano-etal-2025-effective} to medical QA \citep{kim-yoon-2025-questioning}, explicit numerical reasoning remains a critical bottleneck. Prior efforts primarily focus on textual or structured data processing \citep{bahrololloomi-etal-2025-transformer}, largely bypassing complex calculation. Even recent domain-specific models~\citep{chen2024huatuogpto1,lai2025med} often treat mathematical formulas as static text patterns rather than executable logic. This reliance on unverified generation leads to hallucinations in calculation, rendering current systems insufficient for reliable clinical deployment.

%% file: sections/3_method.tex
\section{Method} \label{sec:method}

\subsection{Problem Formulation}
The task of medical mathematical reasoning requires extracting key clinical information from text and performing precise calculations. Formally, given an input $x=(c, q)$, where $c$ denotes the clinical context (e.g., medical records or case descriptions) and $q$ denotes the task instruction (e.g., compute creatinine clearance or adjust drug dosage), the model is expected to generate $y=(f, r, v)$, where $f$ is an explicit formula representation, $r$ denotes the step-by-step reasoning process, and $v \in \mathbb{R}$ is the numerical result. The model parameters are denoted by $\theta$, and the conditional distribution is $\pi_\theta(y \mid x)$.

Following \citet{khandekar2024medcalcbench}, we evaluate floating-point accuracy using a tolerance interval $[L, U] = [v^* - \epsilon, v^* + \epsilon]$ around the ground truth $v^*$, where a prediction is correct if $v \in [L,U]$. Besides, clinical safety imposes a strict transparency constraint: the generated formula f must explicitly align with medical guidelines. This effectively prevents fortuitous approximation, meeting the interpretability standards of high-stakes decision-making. Thus, the overall optimization objective is:
\begin{equation}
    \max_\theta \;\; \mathbb{E}_{x \sim \mathcal{D}, \; y \sim \pi_\theta(\cdot|x)} \big[ R(y; x) \big]
\end{equation}
where $R(y; x)$ denotes the knowledge-guided reward mechanism.

\subsection{Overview of Knowledge-Guided Reward Framework}
We adopt a two-stage training paradigm, as illustrated in Figure~\ref{fig:main}. The model is first initialized via SFT to acquire instruction following and domain knowledge, followed by RL to refine reasoning transparency and numerical precision. During the RL stage, we introduce a knowledge-guided reward framework comprising three synergistic components:
\textbf{\textit{i. Format Reward ($R_f$)}:} Enforces strict structural adherence, requiring explicit formula representation, step-by-step reasoning, and a clear final answer. This penalty mechanism ensures the output remains machine-parsable and auditable.
\textbf{\textit{ii. Knowledge Reward ($R_k$)}:} Verifies whether the generated formula semantically aligns with medical guidelines and input constraints. This component is critical for preventing formula hallucinations and mitigating catastrophic forgetting of domain knowledge.
\textbf{\textit{iii. Answer Reward ($R_a$)}:} Evaluates the numerical fidelity of the result via a hybrid soft-hard mechanism. A \textit{hard constraint} assigns a binary reward based on the clinical tolerance interval $[L, U]$, ensuring safety, while a \textit{soft reward} provides continuous feedback based on the deviation from the ground truth $v^*$, guiding precise convergence.
\begin{equation}
\resizebox{0.89\linewidth}{!}{$
R(y; x) = \alpha \cdot R_{\text{f}}(y; x) + \beta \cdot R_{\text{k}}(y; x) + \gamma \cdot R_{\text{a}}(y; x)     \label{equ:reward}
$}
\end{equation}
where $\alpha, \beta, \gamma \geq 0$ balance the contributions of the three reward components.

\subsection{Knowledge Verification Reward}
To mitigate formula hallucinations and catastrophic forgetting during RL, we introduce a formula-level verification mechanism. Specifically, the model is required to generate an explicit formula representation $f$. We employ a stronger, frozen language model $\mathcal{V}$ as an external knowledge judge to assess the semantic validity of $f$ against the clinical context $x$. The verifier determines whether $f$ aligns with the task-specific valid formula set $\Phi(x)$, yielding a binary reward:
\begin{equation}
    R_{\text{k}}(f,x) =
        \begin{cases}
        1.0, & \text{if } f \text{ is validated by } \mathcal{V}, \\[6pt]
        -1.0, & \text{otherwise}.
        \end{cases}
\end{equation}
This mechanism acts as a strong logical prior, constraining the policy to recall accurate medical knowledge and ensuring that the reasoning process remains transparent and clinically grounded.

\subsection{Hybrid Hard-Soft Reward}






Medical applications impose strict safety requirements, and relying solely on coarse \textit{tolerance interval} rewards is insufficient to achieve consistent improvements in numerical accuracy. To address this, we propose a hybrid soft-hard reward mechanism that safeguards clinical safety while enhancing precision.

\paragraph{Hard Constraint.} 
To strictly enforce clinical safety boundaries, we implement a binary reward mechanism based on the tolerance interval $[L, U]$. The hard reward function is formulated as:
\begin{equation}
    R_{\text{hard}}(v; x) =
        \begin{cases}
        r^+, & v \in [L, U], \\[6pt]
        -r^-, & v \notin [L, U].
        \end{cases}
\end{equation}
where $r^+, r^- > 0$ denote the reward magnitude for compliance and the penalty severity for violation, respectively. This discrete design ensures that adhering to the clinically acceptable range acts as a non-negotiable prerequisite, filtering out unsafe predictions before fine-grained optimization.

\paragraph{Soft Reward.}
To encourage precise convergence towards the ground truth $v^*$, we complement the binary hard constraint with a continuous soft reward mechanism. Formally, it is defined as:
\begin{equation}
    R_{\text{soft}}(v; v^*) = \exp\!\left(- \frac{|v - v^*|}{\tau}\right)
\end{equation}
where $\tau > 0$ is a temperature hyperparameter controlling the sensitivity to numerical deviation. This formulation provides a dense gradient signal, assigning higher values as the prediction approaches the exact truth.
Consequently, the total answer reward integrates these complementary components:
\begin{equation}
    R_{\text{a}}(y; x) = R_{\text{hard}}(v; x) + R_{\text{soft}}(v; v^*)
\end{equation}
This hybrid design establishes a coarse-to-fine optimization landscape: $R_{\text{hard}}$ enforces the non-negotiable safety boundaries, while $R_{\text{soft}}$ provides the fine-grained guidance necessary for minimizing numerical error, ensuring the model acts reliably within the clinically valid range.

\subsection{Training Procedure}
The overall training procedure consists of two stages: SFT and RL. 
Table~\ref{tab:instruction} presents the training template used across experiments, and the detailed optimization procedure is summarized in Algorithm~\ref{alg:training}.

\paragraph{Stage I: SFT.} In the first stage, we train the model on medical mathematical reasoning samples $(x, y^*)$ using standard teacher forcing, minimizing the cross-entropy loss:
\begin{equation}
\resizebox{0.89\linewidth}{!}{$
    \mathcal{L}_{\text{SFT}}(\theta) = - \mathbb{E}{(x, y^*)}\!\left[ \sum_t \log \pi_\theta(y_t^* \mid y_{<t}^*, x) \right]
$}
\end{equation}
where $\pi_\theta$ denotes the policy parameterized by $\theta$. This stage is crucial for aligning the model's output with the target format $(f, r, v)$ and injecting the preliminary medical knowledge necessary for stable reinforcement learning.

\paragraph{Stage II: RL via GRPO.} 
Building upon the SFT initialization, we employ Group Relative Policy Optimization (GRPO)~\citep{guo2025deepseek} to refine the model's adherence to safety and precision constraints. We select GRPO for its training stability and memory efficiency by normalizing advantages within a sampled group. Concretely, for each input $x$, the model samples a set of candidate outputs  $\{y_i\}_{i=1}^K$under policy $\pi_\theta$. Each candidate is evaluated with the proposed reward function, as defined in Eq.~\eqref{equ:reward}. The policy parameters are then updated according to the GRPO gradient estimator:
\begin{equation}
\label{eq:grpo}
\resizebox{0.89\linewidth}{!}{$ 
    \begin{aligned}
        \mathcal{J}_{\text{GRPO}}(\theta) &= \mathbb{E}_{x \sim \mathcal{D}, \{y_i\} \sim \pi_{\theta}} \Bigg[ \frac{1}{G} \sum_{i=1}^G \bigg( \\
        & \quad \min \left( P \cdot A_i, C \cdot A_i \right) - \beta \mathbb{D}_{\text{KL}} \left(\pi_\theta || \pi_{\text{ref}} \right) \bigg) \Bigg]
    \end{aligned}
$}
\end{equation}
\begin{equation}
    \resizebox{0.89\linewidth}{!}{$
    P=\frac{\pi_\theta(y_i|x)}{\pi_{\theta_\text{old}}(y_i|x)}, C=\operatorname{clip}\left(\frac{\pi_\theta(y_i|x)}{\pi_{\theta_\text{old}}(y_i|x)},1-\epsilon,1+\epsilon\right)
    $}
\end{equation}
where $A$ is the advantage value, $x$ is the input, $y_i$ is the response generated by LLMs, and $\pi_\theta(y|x)=\prod_{j=1}^{|y_i|}\pi_\theta(y_{i,j}|x, y_{i, <j})$ is the generation probability of the response $y_i$ under policy $\pi$.

%% file: sections/4_experiment.tex
\section{Experiments} \label{sec:experiments}

\subsection{Datasets}

We evaluate our method on MedCalc-Bench~\citep{khandekar2024medcalcbench}, the premier large-scale benchmark tailored for medical mathematical reasoning. The dataset encompasses 55 clinical tasks categorized into equation-based computation (36 tasks) and rule-based reasoning (19 tasks). Each instance is richly annotated with clinical case descriptions, gold-standard answers, and step-by-step rationales, enabling a holistic evaluation of formula recall, parameter extraction, and arithmetic accuracy. Detailed statistics are provided in Table \ref{tab:dataset}. The training set comprises 9,765 instances across 38 subtasks. Crucially, the test set (1,048 instances) spans 57 subtasks, introducing specific diagnostic criteria and laboratory computations absent from the training phase. This setup rigorously evaluates the model's cross-task generalization capability in novel clinical contexts.

\subsection{Baselines}

We evaluate our approach against a diverse suite of baselines categorized into three groups: \begin{itemize}[itemsep=2pt,topsep=2pt,parsep=2pt,leftmargin=*]
\item \textbf{Proprietary Baselines:} We include GPT-4o~\citep{hurst2024gpt4o} and GPT-3.5-Turbo as references for general-purpose reasoning. Additionally, we evaluate reasoning-specialized models, o1-mini~\citep{jaech2024openai} and DeepSeek-R1~\citep{guo2025deepseek}, to establish upper-bound performance for long-chain inference. 
\item \textbf{Open-Source Baselines:} We select models exhibiting strong mathematical and medical competencies across various scales. These include the DeepSeek-R1-Distill family~\citep{guo2025deepseek}, the Qwen2.5 series~\citep{qwen2.5}, QwQ-32B~\citep{qwq32b}, and the domain-specific HuatuoGPT-o1~\citep{chen2024huatuogpto1}. This selection allows us to benchmark against state-of-the-art open-source capabilities. 
\item \textbf{SFT Baselines:} To rigorously quantify the benefits of our RL framework over standard supervision, we fine-tune Qwen2.5-Instruct (1.5B and 3B) on the MedCalc training set. These variants serve as a direct control group to demonstrate the efficacy of our proposed method in small-scale, domain-specific settings. 
\end{itemize}

\subsection{Implementation Details}
All experiments were conducted on four NVIDIA H800 (80GB) GPUs. We deployed open-source baselines using vLLM, configured with a temperature of 1.0 and a maximum token limit of 2,048, while proprietary models were accessed via official APIs. For our proposed \textsc{MedCalc-R1}, we adopted Qwen2.5-Instruct (1.5B and 3B) as backbones. These efficient scales were specifically chosen to demonstrate that our framework maintains high reasoning robustness while significantly reducing computational overhead. More experiment setting details see Appendix \ref{sec:setting}.


\input{tables/main_results}
\subsection{Main Results}


Table \ref{tab:main_results} presents the main experimental results on the MedCalc-Bench dataset. Overall, closed-source models remain dominant, with DeepSeek-R1 and o1-mini representing the current upper bound of general-purpose reasoning capabilities. In contrast, existing open-source models show limited performance on medical mathematical reasoning. Even the large-scale Qwen2.5-32B-Instruct achieves only 39.03 on average, far below the closed-source counterparts.
By comparison, our proposed \textsc{MedCalc-R1} demonstrates significant advantages at equal or even smaller parameter scales. Specifically, \textsc{MedCalc-R1}$_{\text{1.5B}}$ achieves an average score of 42.36, already surpassing Qwen2.5-32B-Instruct. Furthermore, \textsc{MedCalc-R1}$_{\text{3B}}$ sets a new SOTA among open-weight models with a score of 51.34, substantially narrowing the gap with closed-source counterparts. These results indicate that the proposed optimization framework effectively enhances medical mathematical reasoning even with relatively compact models.

In \textit{equation-based tasks}, our method yields substantial gains in high-risk scenarios. For Dosage calculation, \textsc{MedCalc-R1}$_{\text{1.5B}}$ and \textsc{MedCalc-R1}$_{\text{3B}}$ achieve 90.00 and 87.50, respectively, outperforming o1-mini by a large margin. This highlights the value of our reward design in high-risk clinical scenarios. Similar trends are observed in Physical and Lab tasks, where \textsc{MedCalc-R1}$_{\text{3B}}$ surpasses the SFT baseline by over 10 points. While small models remain unstable on Date calculations, performance improves notably with scale.
In \textit{rule-based tasks}, the model excels in Diagnosis, with the 1.5B model reaching 73.33, suggesting that explicit reasoning steps enhance the interpretability of clinical decision rules. However, performance on Severity remains limited. This task involves complex hierarchical constraints and discrete rule enumeration, indicating that current reward mechanisms may require augmentation with external knowledge bases to handle such structural logical dependencies.

\begin{table}
  \vspace{-10pt} 
  \centering
  \resizebox{\linewidth}{!}{
  \begin{tabular}{lccc}
  \toprule
    \multicolumn{1}{c}{\textbf{Model}} & \textbf{Equation} & \textbf{Rule-based} & \textbf{Avg} \\  \midrule
    \textsc{MedCalc-R1}$_\text{3B}$                         & 60.12             & 35.96                    &    51.34          \\
    \quad w/o KnoRe                 &   53.52$_{\downarrow \textbf{6.60}}$                &   35.17$_{\downarrow \textbf{0.79}}$                  &    46.85$_{\downarrow \textbf{4.49}}$          \\
    \quad w/o HyRe                    &   54.87$_{\downarrow \textbf{5.25}}$                &   31.23$_{\downarrow \textbf{4.73}}$                  &    46.28$_{\downarrow \textbf{5.06}}$          \\
    \quad w/o Both                             &  43.02$_{\downarrow \textbf{17.10}}$                 & 33.60$_{\downarrow \textbf{2.36}}$                    &   39.60$_{\downarrow \textbf{11.74}}$                                 \\
\bottomrule
  \end{tabular}
  }
  \vspace{-2mm}
  \caption{Ablation results on MedCalc-Bench, showing the performance impact of removing the knowledge reward (KnoRe) and hybrid reward (HyRe) mechanisms.}
  \label{tab:ablation}
\end{table}

\subsection{Ablation Study}

Table \ref{tab:ablation} details the component-wise contribution of our reward framework. Overall, the dual-reward mechanism proves essential: removing the knowledge reward or the hybrid reward degrades average performance by 4.49\% and 5.06\%, respectively, while eliminating both precipitates a steep decline of 11.74\%. This confirms that the two components are not merely additive but complementary.

\paragraph{\textit{Task-specific sensitivity.}} Further analysis reveals distinct dependency patterns. Equation-based tasks exhibit high sensitivity to the knowledge reward. This suggests that explicit formula verification effectively mitigates formula hallucinations and prevents catastrophic forgetting of domain knowledge. Conversely, rule-based tasks benefit predominantly from the hybrid reward. This indicates that the soft-hard constraint design is critical for aligning discrete decision boundaries with clinical safety thresholds. For a deeper empirical investigation into how the hybrid reward mitigates policy stagnation compared to standard tolerance-based methods, please refer to  Appendix \ref{appd:tole}.

\paragraph{\textit{Synergistic necessity.}} Notably, deploying the Knowledge Reward in isolation yields performance inferior even to the baseline (where both are removed). This implies that enforcing formula correctness without concurrent numerical precision constraints (provided by the hybrid reward) induces optimization instability. Thus, the joint application of both rewards is non-negotiable for reliable medical reasoning.


\begin{figure}[]
\centering
\includegraphics[width=0.9\linewidth]{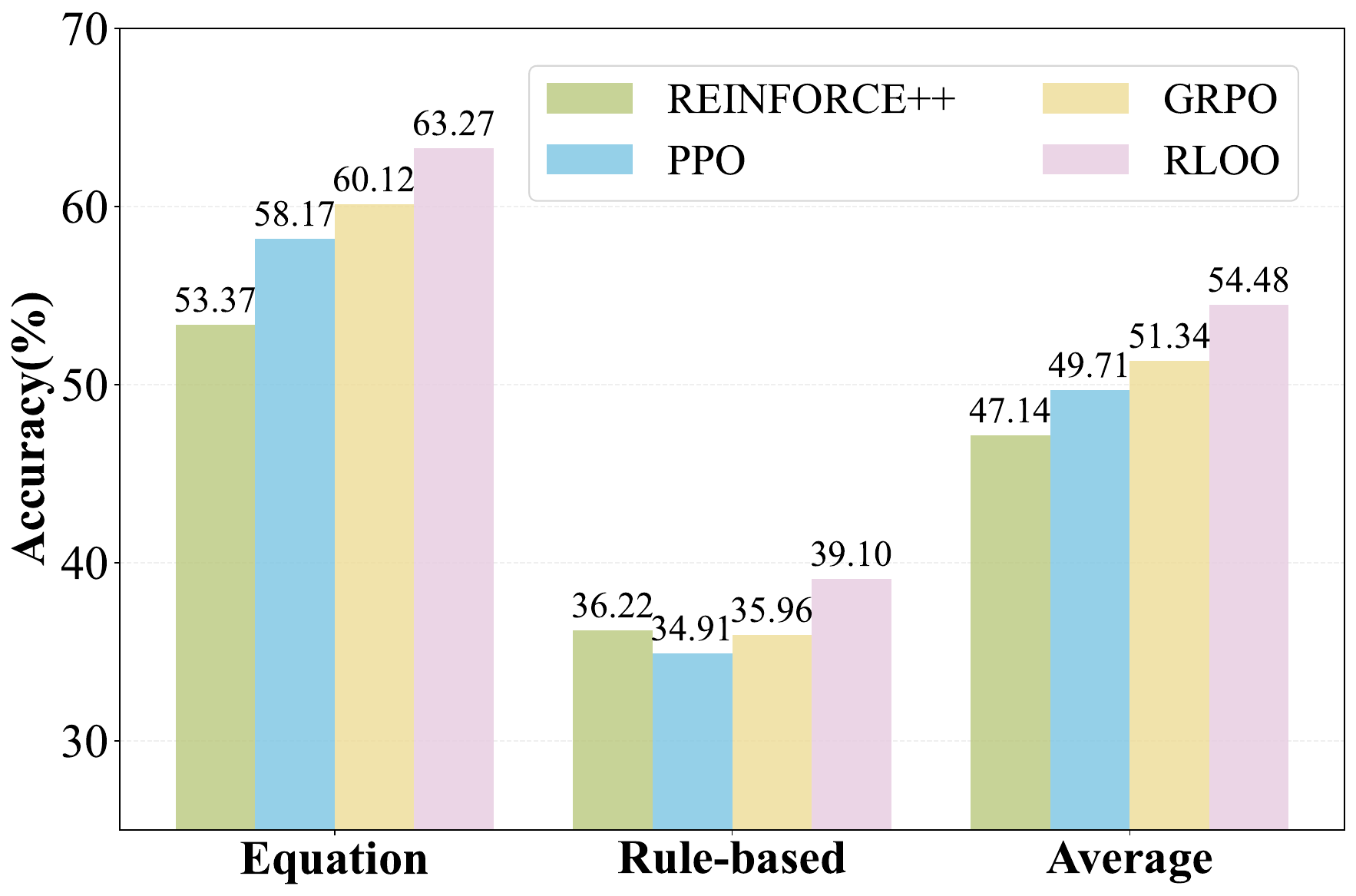}
\vspace{-3mm}
\caption{Performance comparison under different reinforcement learning algorithms.}
\label{fig:rl_diff}
\end{figure}

\subsection{Further Analysis}

\paragraph{\textit{Group-relative optimization strategies prove significantly more effective.}}
As illustrated in Figure \ref{fig:rl_diff}, RLOO~\cite{ahmadian-etal-2024-back} achieves the strongest overall performance, leveraging leave-one-out baselines to effectively reduce gradient variance and refine arithmetic precision in equation-based tasks. GRPO ranks second, consistently surpassing PPO~\cite{schulman2017proximal} and REINFORCE++~\cite{hu2025reinforce} by utilizing group-wise normalization to stabilize training dynamics. In contrast, while PPO provides consistent but lower-ceiling results, REINFORCE++ suffers from significant volatility in equation reasoning due to aggressive exploration, despite exhibiting potential in rule-based tasks. These findings underscore that variance reduction mechanisms are more critical than proximal constraints for optimizing high-precision clinical reasoning.






\begin{figure}[]
\centering
\includegraphics[width=0.9\linewidth]{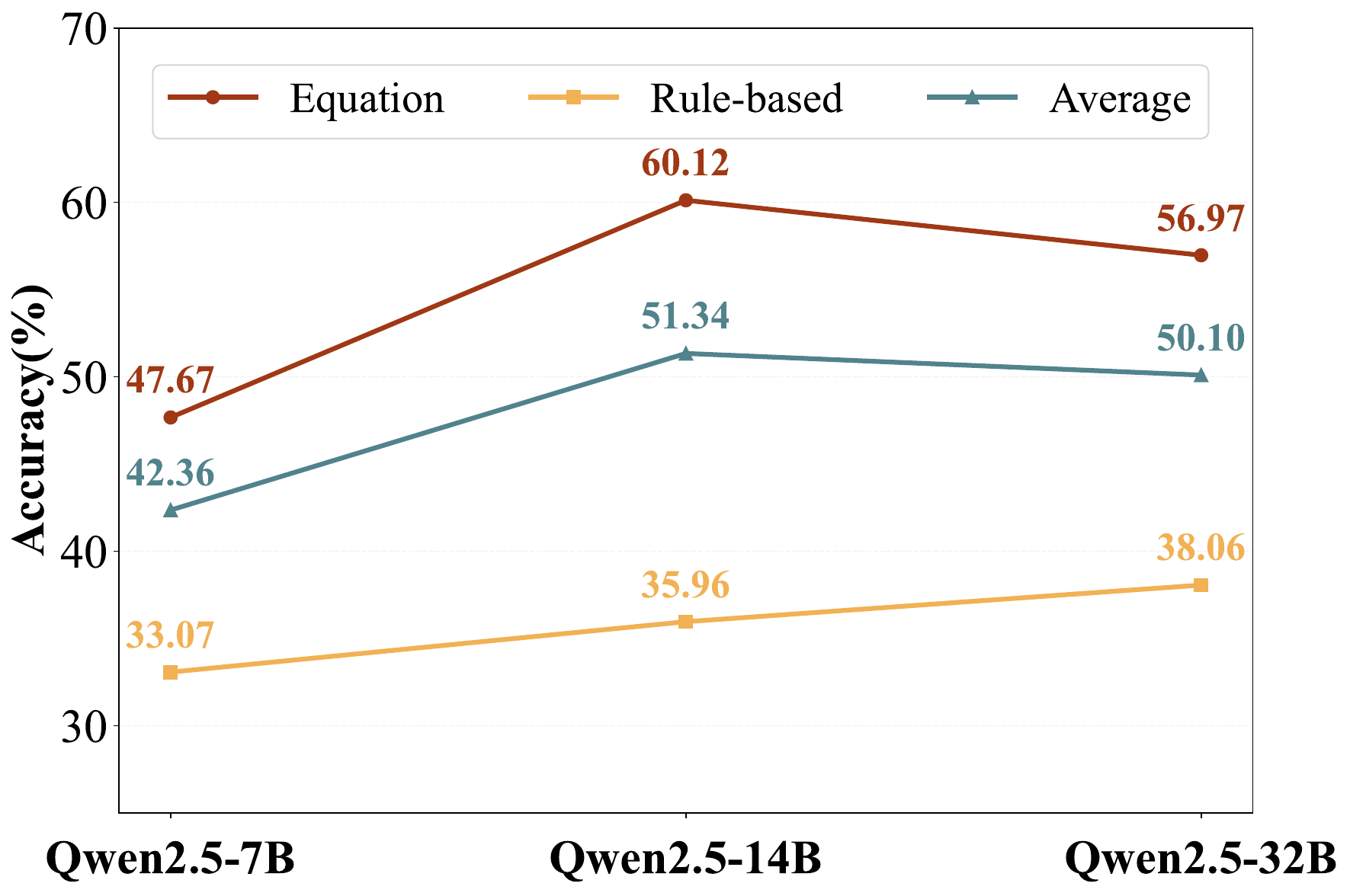}
\vspace{-3mm}
\caption{Impact of verifier capacity for model performance.}
\label{fig:reward}
\end{figure}

\paragraph{\textit{Verifier scaling yields non-linear performance gains.}}
We tested Qwen2.5 models of different scales (7B, 14B, 32B) as knowledge verification models. As shown in Figure~\ref{fig:reward}, while the 7B model suffers from inadequate calibration, the 32B model paradoxically underperforms the 14B variant on average. Task-level decomposition clarifies this divergence: equation-based tasks exhibit high sensitivity to verifier size, improving by 12.45\% from 7B to 14B but degrading at 32B. We attribute this regression to the larger model's excessive strictness, which imposes overly rigid constraints and induces reward sparsity. Conversely, rule-based tasks benefit monotonically from increased scale, confirming that larger capacities are essential for capturing fine-grained decision boundaries. Thus, 14B provides the best trade-off, offering sufficient precision without stifling exploration through hyper-correction.

\begin{figure}[]
\centering
\includegraphics[width=0.9\linewidth]{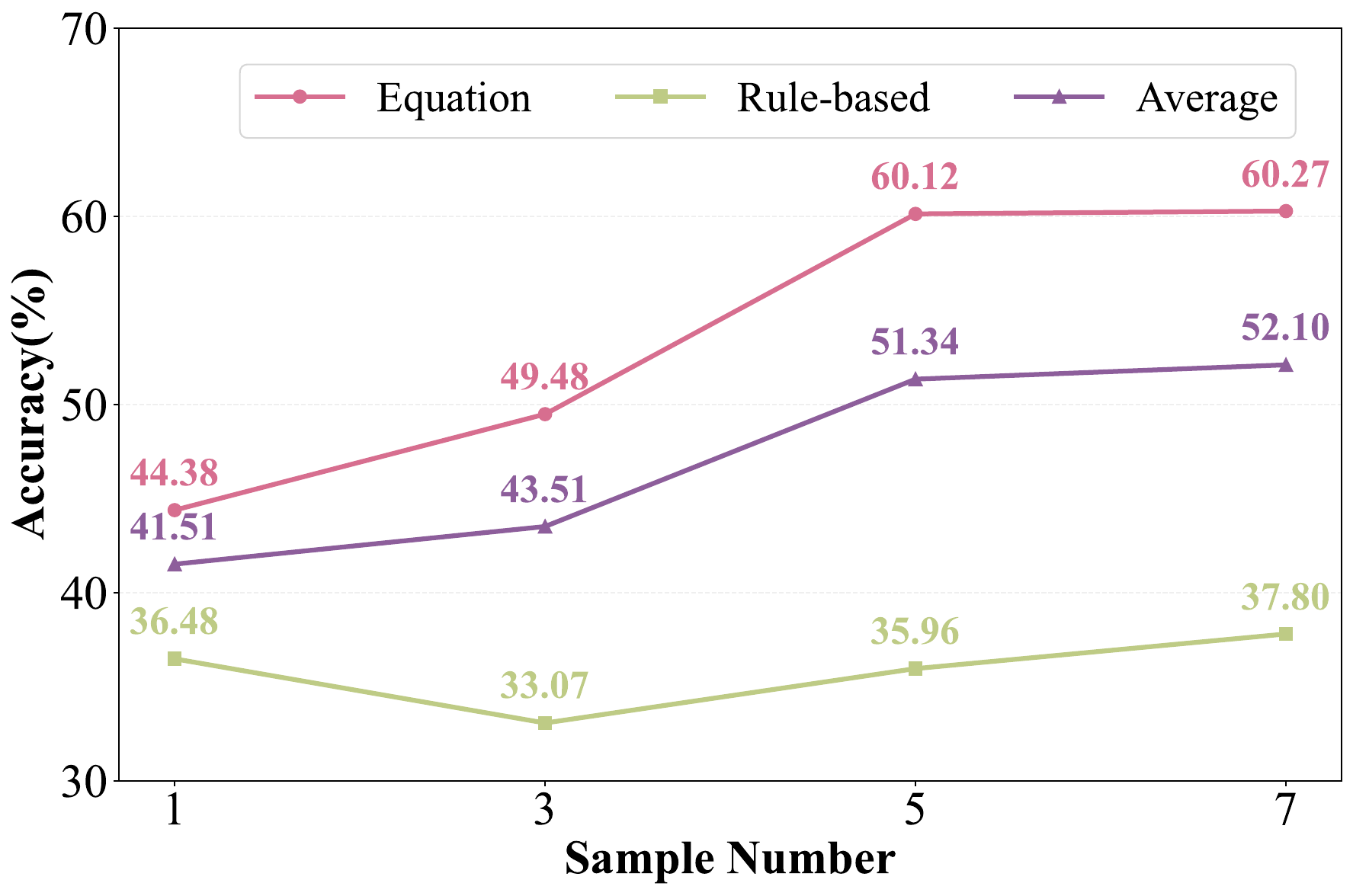}
\vspace{-3mm}
\caption{Effect of rollout sample size under the GRPO framework.}
\label{fig:sample}
\end{figure}


\paragraph{\textit{Increasing the number of rollouts yields diminishing marginal returns.}} We experimented with varying rollouts, as illustrated in Figure~\ref{fig:sample}. While small sample rollouts sizes suffer from high-variance baseline estimation, increasing rollouts beyond five leads to performance saturation where additional computational costs outweigh the marginal accuracy gains. Task-level analysis reveals that equation-based tasks are particularly sensitive to group size. This sensitivity arises because larger samples provide a more stable baseline for advantage estimation, which is critical for refining precise numerical reasoning. In contrast, rule-based tasks exhibit earlier saturation but rely on sufficient samples to smooth out training fluctuations. Consequently, we set rollouts to 5 as the standard configuration to ensure robust convergence without incurring excessive overhead.





\begin{figure}[]
\centering
\includegraphics[width=0.9\linewidth]{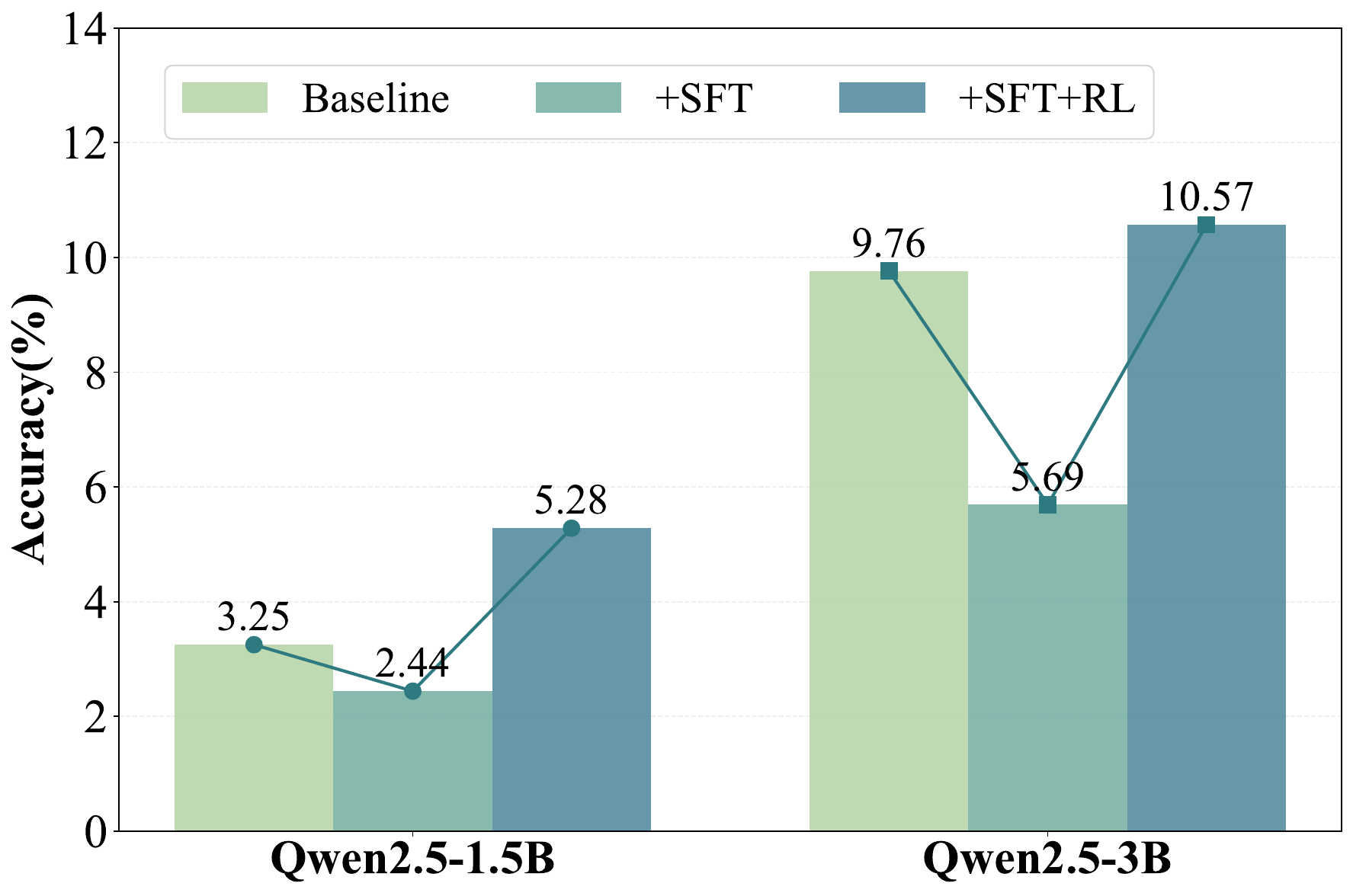}
\vspace{-3mm}
\caption{Cross-task generalization on unseen tasks.}
\label{fig:generalization}
\end{figure}

\subsection{Generalization Analysis}

\paragraph{\textit{RL significantly enhances out-of-distribution (OOD) generalization.}}
To evaluate cross-task generalization, we tested different training methods on unseen tasks. As shown in Figure~\ref{fig:generalization}, the SFT+RL configuration consistently outperforms both base and SFT-only variants across all scales. Notably, relying solely on SFT degrades performance on unseen tasks, particularly in larger models. This suggests that SFT tends to overfit the training distribution by memorizing superficial heuristics rather than learning robust reasoning. Mechanistically, our RL framework introduces inductive biases that are independent of the data distribution. The formula verification reward enforces structural correctness to prevent knowledge forgetting, while the hybrid soft-hard reward aligns numerical outputs with rigorous clinical constraints. These process-oriented mechanisms compel the model to internalize medical calculation logic rather than surface data patterns, yielding superior transferability to novel clinical scenarios.

\begin{figure}[]
\centering
\includegraphics[width=0.9\linewidth]{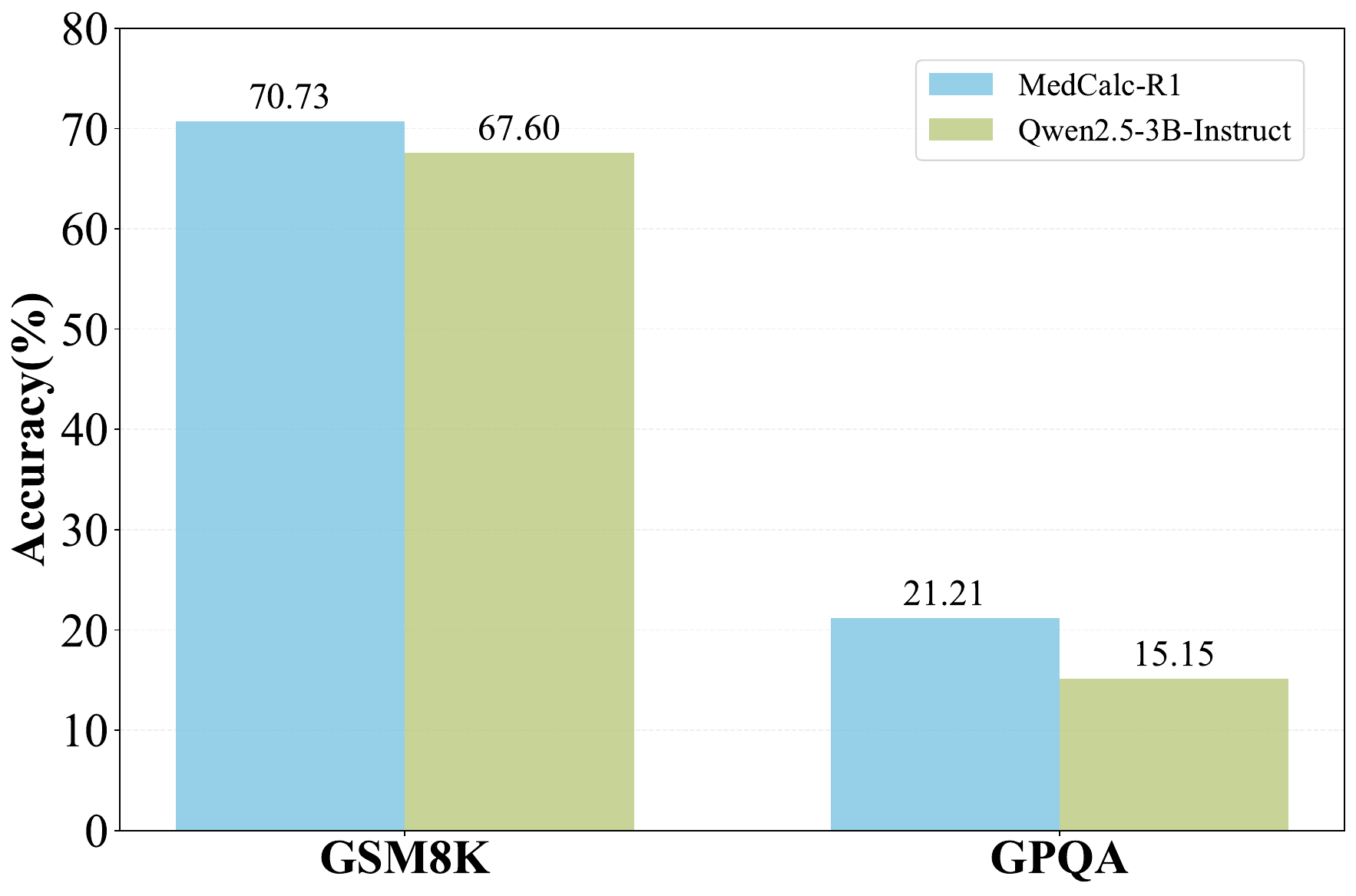}
\vspace{-3mm}
\caption{Generalization ability in open-domain numerical reasoning tasks.}
\label{fig:generalization-open}
\end{figure}

\paragraph{\textit{Knowledge-constraint framework promotes more stable intermediate reasoning and stronger constraint awareness.}}
To evaluate the cross-domain generalization ability of MedCalc-R1, we further conduct experiments on two open-domain numerical reasoning benchmarks, GSM8K\cite{cobbe2021training} and GPQA\cite{rein2024gpqa}, as shown in Figure~\ref{fig:generalization-open}. MedCalc-R1 consistently outperforms the original Qwen2.5-3B-Instruct model, yielding improvements of 3.13\% on GSM8K and 6.06\% on GPQA. The larger gain on GPQA, a more challenging benchmark, indicates that our method improves not only medical numerical reasoning but also general reasoning ability in more open and complex problem settings. 
This result suggests that the proposed reinforcement learning framework does not simply specialize the model to medical tasks. Instead, the formula-level knowledge verification reward and hybrid soft-hard constraints encourage more stable intermediate reasoning and stronger constraint awareness. Since these mechanisms rely on process-level verification rather than medical knowledge alone, MedCalc-R1 shows promising transferability to other numerical reasoning scenarios involving explicit rules, computational procedures, and reliability requirements.

%% file: tables/main_results.tex
\begin{table*}[]
\centering
\resizebox{\linewidth}{!}{
\begin{tabular}{ccccccccc}
\toprule
\multirow{2}{*}{\textbf{Model}} & \multicolumn{4}{c}{\textbf{Equation}} & \multicolumn{3}{c}{\textbf{Rule-based}} & \multirow{2}{*}{\textbf{Avg}} \\  \cmidrule(lr){2-5} \cmidrule(lr){6-8}
                       & \textbf{Lab}  & \textbf{Physical} & \textbf{Date} & \textbf{Dosage} & \textbf{Risk}      & \textbf{Severity}     & \textbf{Diagnosis}     &                      \\ \midrule
\rowcolor[gray]{.92}
GPT-3.5-Trubo                           &  21.47   &   23.24    &  8.33   &   0.00    &   21.16    &   13.75    &   25.00    &  19.85     \\
\rowcolor[gray]{.92}
GPT-4o                                  &  45.71   &   43.98    &  50.00  &   42.50   &   24.90    &   38.75    &   50.00    &  40.36     \\
\rowcolor[gray]{.92}
o1-mini                                 &  73.01   &   84.23    &  48.33  &   40.00   &   64.73    &   52.50    &   45.00    &  67.84     \\
\rowcolor[gray]{.92}
DeepSeek-R1                             &  67.48   &   93.36    &  66.67  &   57.50   &   71.78    &   56.25    &   81.67    &  73.95     \\ \midrule
DeepSeek-R1-Distill-Qwen-1.5B           &  1.53    &   6.64     &  1.67   &   2.50    &   2.07     &   1.25     &   1.67     &  2.86      \\
Qwen2.5-1.5B-Instruct                   &  1.53    &   8.10     &  1.67   &   5.00    &   5.39     &   2.50     &   12.70    &  4.82      \\
DeepSeek-R1-Distill-Qwen-7B             &  5.52    &   21.26    &  11.67  &   0.00    &   10.37    &   2.50     &   16.67    &  10.78     \\
Qwen2.5-3B-Instruct                     &  4.29    &   13.36    &  5.00   &   0.00    &   17.84    &   16.25    &   41.27    &  12.49     \\
DeepSeek-R1-Distill-Qwen-14B            &  9.82    &   26.97    &  \textbf{46.67}  &   5.00    &   18.26    &   13.75    &   43.33    &  19.85     \\
Qwen2.5-7B-Instruct                     &  19.63   &   29.55    &  18.33  &   5.00    &   19.92    &   \underline{23.75}    &   47.63    &  23.37     \\
HuatuoGPT-o1$^\clubsuit$                &  21.17   &   33.61    &  13.33  &   7.50    &   24.48    &   16.25    &   40.00    &  24.52     \\ 
DeepSeek-R1-Distill-Qwen-32B            &  20.25   &   34.02    &  30.00  &   10.00   &   19.92    &   20.00    &   45.00    &  24.90     \\
Qwen2.5-14B-Instruct                    &  26.99   &   35.68    &  26.67  &   2.50    &   29.88    &   \underline{23.75}    &   38.33    &  29.10     \\
QwQ-32B                                 &  26.99   &   40.25    &  38.33  &   15.00   &   24.90    &   \textbf{32.50}    &   55.00    &  31.77     \\
Qwen2.5-32B-Instruct                    &  34.05   &   51.87    &  \textbf{46.67}  &   10.00   &   \underline{33.61}    &   \textbf{32.50}    &   56.67    &  39.03     \\
\midrule
Qwen2.5-1.5B-Instruct$^\spadesuit$      &  25.46   &   54.35    &  28.33  &   \textbf{32.50}   &   26.97    &   10.00    &   60.00    &  33.68     \\
Qwen2.5-3B-Instruct$^\spadesuit$        &  34.97   &  \underline{70.54}     &  38.33  &   17.50   &   29.05    &   8.75     &   58.33    &  40.65     \\
\midrule
\textsc{MedCalc-R1}$_{\text{1.5B}}$                     &  \underline{36.50}   &   64.73    &  26.66  &   \textbf{90.00}   &   26.56    &   11.24    &   \textbf{73.33}    &  \underline{42.36}  \\
\textsc{MedCalc-R1}$_{\text{3B}}$                       &  \textbf{43.25}   &   \textbf{82.15}    &  \underline{45.00}  &   \underline{87.50}   &   \textbf{36.93}    &   8.75     &   \underline{68.33}    &  \textbf{51.34}  \\
                       
\bottomrule
\end{tabular}
}
\vspace{-2mm}
\caption{Main results on the MedCalc-Bench dataset. Rows shaded in gray represent reference performance from advanced proprietary or open-source models. Among the remaining comparable methods, \textbf{bold} and \underline{underlined} values denote the best and second-best results, respectively.
$^\spadesuit$ marks models fine-tuned on medical numerical reasoning tasks, while $^\clubsuit$ refers to domain-specific medical LLMs.
}
\label{tab:main_results}
\end{table*}

%% file: sections/5_conclusion.tex
\section{Conclusion}


In this work, we presented \textsc{MedCalc-R1}, a knowledge-guided reinforcement learning framework designed to effectively mitigate calculation hallucinations and resolve the stability-precision trade-off inherent in tolerance-based evaluation. Mechanistically, we integrate a formula-level verification module that utilizes an external judge to enforce semantic consistency, thereby curbing logical hallucinations. This is coupled with a hybrid soft-hard reward strategy, where the hard constraint guarantees predictions fall within the acceptable error interval, while the soft reward provides continuous feedback to drive fine-grained numerical precision. Extensive experiments on MedCalc-Bench demonstrate that \textsc{MedCalc-R1} significantly outperforms state-of-the-art open-source baselines, enabling parameter-efficient models (1.5B/3B) to achieve reasoning robustness comparable to larger counterparts. Furthermore, our framework exhibits strong cross-task generalization, validating the importance of enforcing structural correctness over superficial heuristic matching. 



%% file: sections/limitations.tex
\section*{Limitations}


In this study, we introduced \textsc{MedCalc-R1}, a knowledge-guided reinforcement learning framework that significantly enhances the reliability and numerical precision of clinical reasoning through a hybrid reward mechanism. Despite these advancements, our work remains subject to three primary limitations. First, the evaluation is constrained by the scarcity of specialized benchmarks. We validated our method exclusively on MedCalc-Bench, as it is currently the only dataset containing both calculation traces and safety intervals; the future availability of diverse datasets will allow for more rigorous generalizability testing. Second, our framework is confined to textual modalities. Since real-world clinical diagnosis often necessitates synthesizing multimodal information (e.g., medical imaging and vital signs), extending \textsc{MedCalc-R1} to multimodal contexts remains a critical direction. Third, the knowledge verification module relies on a reference LLM, which may encounter bottlenecks when validating highly complex or implicit medical logic. While effective for standard formulas, future iterations could be further robustified by integrating structured medical knowledge graphs (KGs) to handle intricate hierarchical constraints more precisely.

%% file: sections/ethics_statement.tex
\section*{Ethics Statement}
This work focuses on medical mathematical reasoning using LLMs. All experiments are conducted on the MedCalc-Bench~\cite{khandekar2024medcalcbench} dataset, which consists of publicly available medical case descriptions without any personally identifiable information (PII). We have carefully ensured compliance with privacy and data protection regulations, and no sensitive or private information is disclosed in this study.

We acknowledge the potential societal impact of applying LLMs in clinical domains. While our approach improves accuracy, interpretability, and safety in numerical reasoning, it is not intended for direct clinical deployment without professional oversight. Instead, the proposed framework should be viewed as a research contribution toward safer medical AI. All authors affirm adherence to the Code of Ethics throughout the research and submission process.

%% file: sections/appendix.tex
\clearpage
\appendix



\begin{figure*}[!t] 
    \centering
    \begin{minipage}[b]{0.62\textwidth}
        \centering
        \includegraphics[width=\textwidth]{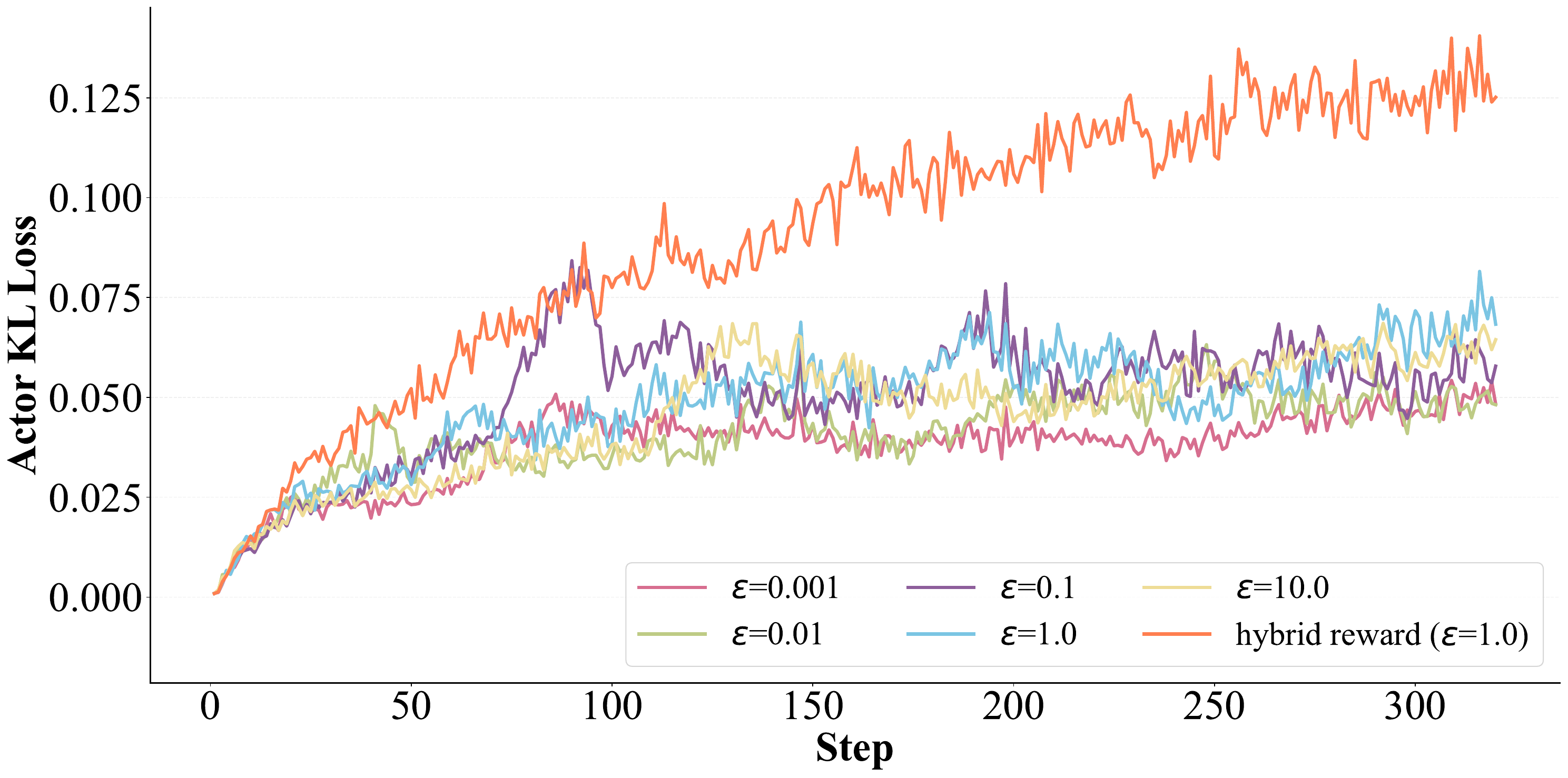}
        \vspace{-8mm}
        \caption{Sensitivity analysis of tolerance thresholds and hybrid reward schema.}
        \label{fig:sensitivity}
    \end{minipage}
    \hfill 
    \begin{minipage}[b]{0.35\textwidth}
        \centering
        \begin{small} 
        \begin{tabular}{cc}
        \toprule
        \textbf{Model}       & \textbf{Accuracy} \\  \midrule
        $\epsilon=0.001$         & 22.45          \\
        $\epsilon=0.01$          & 26.27         \\
        $\epsilon=0.1$           & 24.68         \\
        $\epsilon=1.0$           & 25.00          \\  
        $\epsilon=10.0$          & 23.25          \\ \midrule
        \textbf{Hybrid Reward ($\epsilon=1.0$)}            & \textbf{37.90}          \\ \bottomrule
        \end{tabular}
        \end{small}
        \captionof{table}{Results of different tolerance thresholds. The hybrid reward schema consistently surpasses static tolerance baselines, validating the effectiveness of the coarse-to-fine optimization strategy.}
        \label{tab:sensitivity}
    \end{minipage}
\end{figure*}

\begin{table}[t]
\centering
\resizebox{\linewidth}{!}{
\begin{tabular}{llcccc}
\toprule
\multicolumn{1}{c}{\multirow{2}{*}{\textbf{Task Type}}} & \multicolumn{1}{c}{\multirow{2}{*}{\textbf{Subtask Type}}} & \multicolumn{2}{c}{\textbf{Train Dataset}} & \multicolumn{2}{c}{\textbf{Test Dataset}} \\ \cmidrule(lr){3-4} \cmidrule(lr){5-6}
\multicolumn{1}{c}{}                           & \multicolumn{1}{c}{}                              & \#\textbf{Task}             &   \#\textbf{Inst. }    & \#\textbf{Task}         & \#\textbf{Inst.}         \\   \midrule
\multirow{4}{*}{\textbf{Equation-based}}                & Lab             &    16        &   2,931     &     19      &   326              \\
                                               & Physical        &    12        &   4,804     &     13      &   241             \\
                                               & Date            &    3         &   240       &     3       &   60              \\
                                               & Dosage          &    2         &   151       &     2       &   40             \\ \midrule
\multirow{3}{*}{\textbf{Rule-based}}                    & Risk            &    4         &   1,206     &     13      &   241             \\
                                               & Severity        &    1         &   75        &     4       &   80              \\
                                               & Diagnosis       &    3         &   358       &     3       &   60              \\ \midrule
\multicolumn{2}{c}{\textbf{Overall}}                                      &    38        &   9,765     &     57      &   1,048              \\ \bottomrule
\end{tabular}
}
\caption{Dataset statistics of MedCalc-Bench. The benchmark covers diverse equation-based and rule-based tasks.}
\label{tab:dataset}
\end{table}

\section{LLM Usage Statement}
We used LLMs only as auxiliary tools for translation, grammar polishing, and minor wording improvements. No LLMs were involved in research ideation, experiment design, data analysis, or result interpretation. All methodological development, experiments, and scientific contributions were conducted solely by the authors.

\section{Experiment Setting} \label{sec:setting}
During the SFT phase, models were trained with a learning rate of $1\times10^{-5}$ and a batch size of 8, with checkpoints selected at 400 steps to mitigate overfitting. In the subsequent RL phase, we employed the GRPO algorithm with a learning rate of $1\times10^{-6}$, a batch size of 128, and a rollout size of 5. Training was conducted for 5 epochs with a context window of 2,048 tokens. The formula verification reward was computed using Qwen2.5-14B-Instruct. To ensure generality without task-specific tuning, we set the reward weighting coefficients $\alpha$, $\beta$, $\gamma$ to 1.0, while the hard constraint magnitudes are fixed at 2.0 for $r^+$ and 3.0 for $r^-$, with a soft reward temperature $\tau$ of 1.0. Performance is evaluated using Accuracy, measuring the percentage of predictions with correct numerical outcomes.

\section{Analysis of Training Dynamics and Tolerance Sensitivity} \label{appd:tole}



To strictly validate the impact of reward mechanisms on optimization stability, we conducted a controlled analysis using Qwen-2.5-3B-Instruct on all floating-point reasoning tasks. As illustrated in Figure \ref{fig:sensitivity}, tolerance-based baselines exhibit an intrinsic trade-off between reward sparsity and optimization stability. Specifically, overly strict thresholds (e.g., $\epsilon=0.001$) result in severe reward sparsity, causing policy stagnation; as evidenced by the lowest KL trajectory, the model fails to capture sufficient gradient signals and remains tethered to the SFT initialization, leading to suboptimal convergence. Conversely, excessively lenient thresholds introduce stochastic noise, triggering high-variance oscillations that destabilize parameter updates. In contrast, \textsc{MedCalc-R1} resolves this dilemma via a coarse-to-fine optimization strategy. By combining a hard constraint ($\epsilon =1.0$) to anchor predictions within a feasible safety region with a continuous soft reward for fine-grained approximation, our method provides dense feedback that encourages active learning. Notably, while our approach maintains a higher KL divergence than the stagnant baselines, this reflects effective exploration and the successful breaking of local optima rather than instability. Experiment results (Table \ref{tab:sensitivity}) confirm that this mechanism yields a 11.63\% performance improvement over the best tolerance-based baseline, validating the superiority of the hybrid reward scheme.

\section{Generalization to Different Medical Numerical Reasoning}

\begin{figure}[]
\centering
\includegraphics[width=0.95\linewidth]{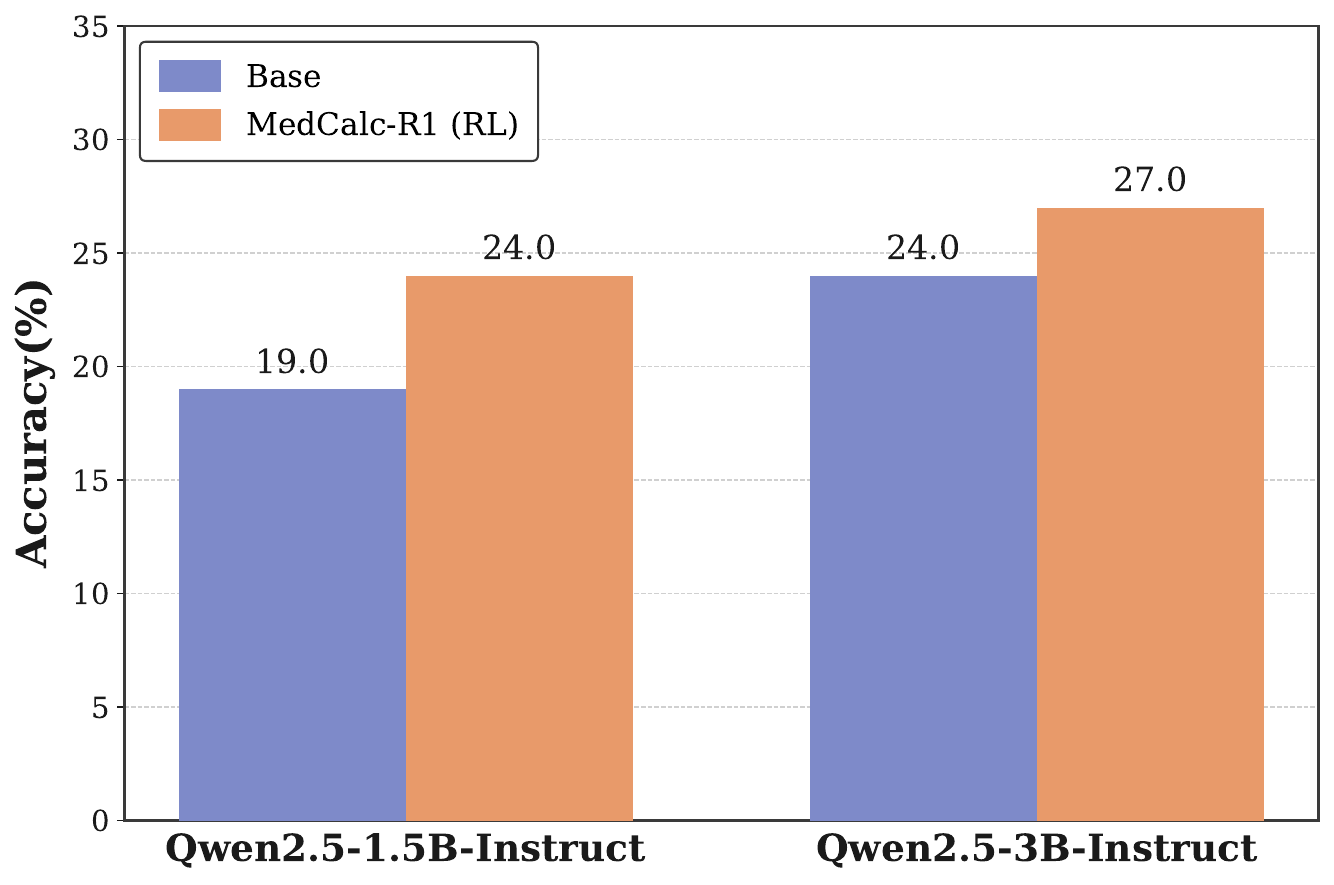}
\vspace{-3mm}
\caption{Generalization results under different medical numerical reasoning.}
\label{fig:generalization-calcqa}
\end{figure}

To further evaluate the generalization ability of MedCalc-R1 across different medical numerical reasoning tasks, we conduct additional experiments on CalcQA~\cite{zhu2025menti}, a rule-based medical numerical reasoning benchmark containing 100 questions. As shown in Figure~\ref{fig:generalization-calcqa}, MedCalc-R1 consistently improves the performance of Qwen2.5-Instruct models at different scales. Specifically, the accuracy of Qwen2.5-1.5B-Instruct increases from 19.0\% to 24.0\%, while that of Qwen2.5-3B-Instruct improves from 24.0\% to 27.0\%. These results indicate that the proposed knowledge-constrained framework not only improves performance on the original medical numerical reasoning tasks, but also transfers effectively to a new rule-based medical calculation benchmark. This demonstrates its generalization ability in formula selection, rule following, and process-constrained computation.

\begin{table}
  \centering
  \resizebox{0.8\linewidth}{!}{
  \begin{tabular}{lc}
    \toprule
    \textbf{Model}       & \textbf{Generalization} \\  \midrule
    PPO         & 8.54          \\
    GRPO        & 10.57         \\
    REINFORCE++ & 10.16         \\
    RLOO        & 9.76          \\  \bottomrule
    \end{tabular}
  }
  \vspace{-2mm}
  \caption{Generalization results under different RL optimizers.}
  \label{tab:generalization}
\end{table}

\section{Generalization Ability of Different RL Optimization Algorithms}
Optimizer fundamentally dictates the policy's capacity for cross-task transfer.
As detailed in Table~\ref{tab:generalization}, GRPO demonstrates superior generalization. Its mechanism of group normalization acts as a dynamic baseline, reducing gradient variance and ensuring stable credit assignment that generalizes across diverse clinical contexts. REINFORCE++, while benefiting from aggressive exploration that improves coverage, suffers from high variance due to the absence of a value baseline, slightly hampering precision on unseen tasks. Conversely, RLOO, though effective on in-distribution data, tends to overfit local sampling artifacts, failing to abstract generalizable reasoning patterns. Finally, PPO exhibits the weakest transferability, suggesting that proximal constraints alone are insufficient for navigating the complex optimization landscape of cross-task medical reasoning.


\begin{table}[]
\centering
\resizebox{\linewidth}{!}{
\begin{tabular}{ccc}
\toprule
\textbf{Model}      & \textbf{GPT-4o Judge} & \textbf{Accuracy} \\  \midrule
Qwen2.5-3B-Instruct &    12.02          &    12.49      \\
Qwen2.5-3B-Instruct$^\spadesuit$ &     53.91         &    40.65      \\
\textsc{MedCalc-R1}          &      \textbf{57.92}        &    \textbf{51.34}      \\  \bottomrule
\end{tabular}
}
\caption{Comparison of formula recall accuracy and overall task performance. $^\spadesuit$ marks models fine-tuned on medical numerical reasoning tasks.}
\label{tab:formula_recall}
\end{table}



\section{Analysis of Formula Recall and Reasoning Consistency}

To further investigate the underlying drivers of \textsc{MedCalc-R1} performance in medical reasoning, we conducted a targeted evaluation of formula recall accuracy. Given that medical reasoning often necessitates the precise retrieval of clinical formulas, we employed GPT-4o as an automated judge to assess the structural and mathematical correctness of formulas invoked during the models' reasoning processes. As illustrated in Table \ref{tab:formula_recall}, several critical insights emerge regarding the transition from knowledge retrieval to logical execution.

The results demonstrate that while SFT model achieves a substantial improvement in formula recall by reaching 53.91 compared to the 12.02 of the base model, a significant \textit{Knowledge-Reasoning Gap} persists. This is evidenced by its comparatively lower final reasoning accuracy of 40.65, suggesting that supervised fine-tuning alone is insufficient to guarantee the correct application of retrieved knowledge within complex logical chains. In contrast, \textsc{MedCalc-R1} not only further optimizes formula recall to 57.92 but also achieves a marked increase in final accuracy to 51.34, representing a 10.69\% absolute gain over the SFT baseline. This indicates that our RL framework effectively disciplines the model’s reasoning trajectory, ensuring higher consistency between formula invocation and step-by-step execution. By mitigating \textit{reasoning hallucinations}, where correct formulas are provided but incorrectly applied, the RL stage successfully bridges the gap between static knowledge recall and active logical inference.

\section{Case Study}

\begin{figure*}[h]
\centering
\includegraphics[width=\linewidth]{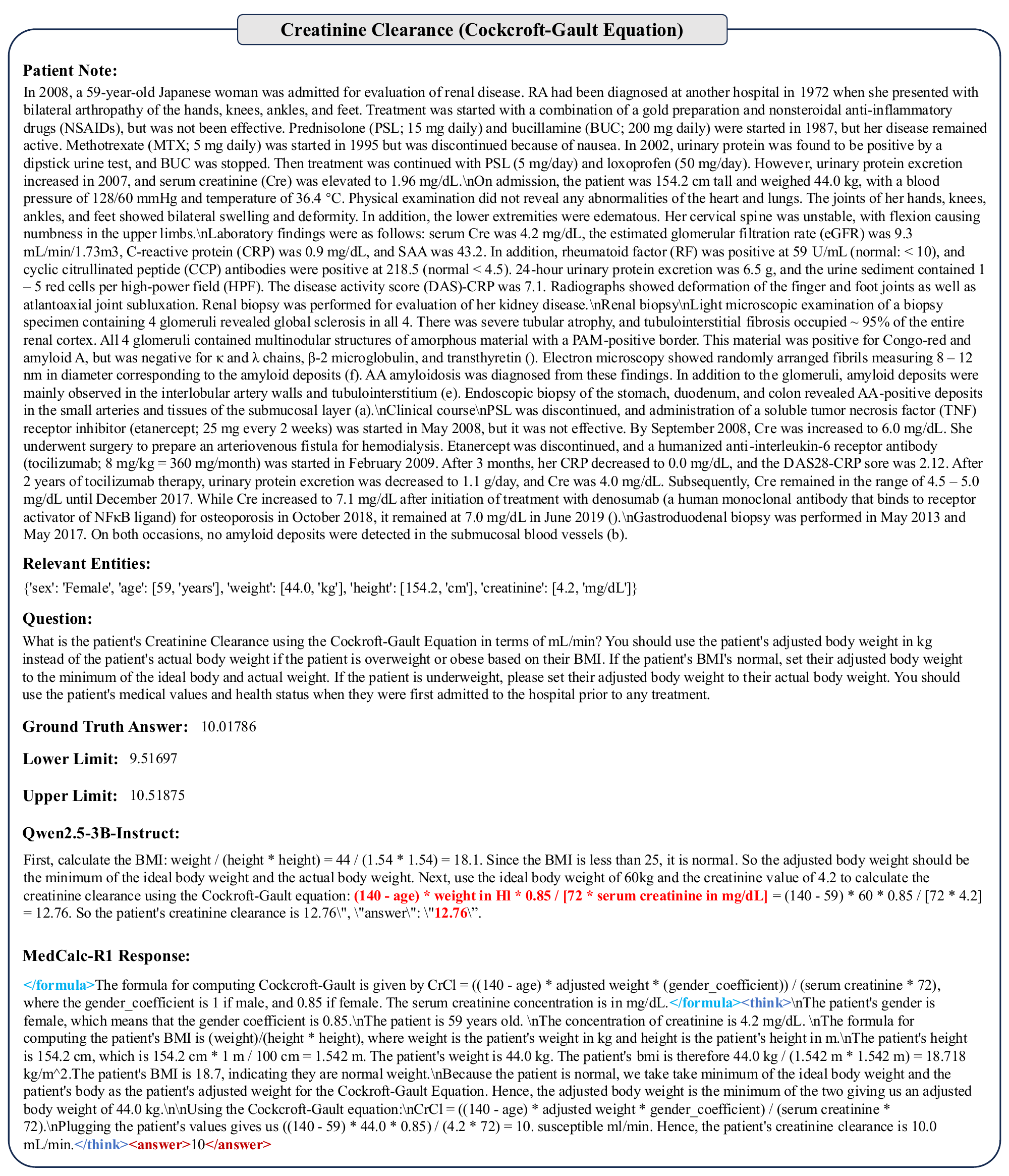}
\caption{Case study on creatinine clearance calculation using the Cockcroft-Gault equation.
}
\label{fig:case1}
\end{figure*}

\begin{figure*}[]
\centering
\includegraphics[width=\linewidth]{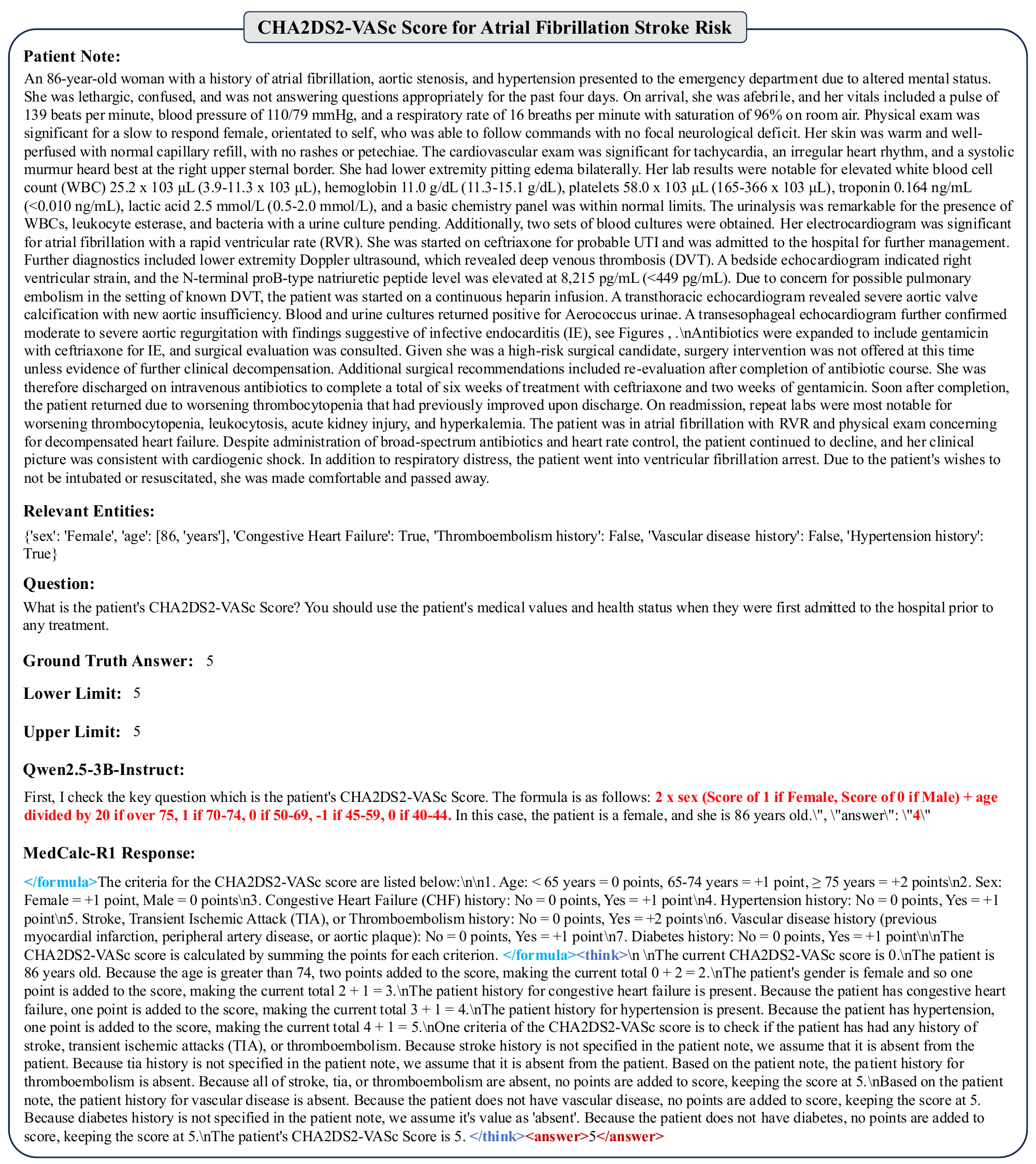}
\caption{Case study on atrial fibrillation stroke risk assessment using the CHA2DS2-VASc score. 
}
\label{fig:case2}
\end{figure*}

To further validate the reasoning ability of the models, we conducted in-depth case analyses, with representative examples shown in Figures \ref{fig:case1} and \ref{fig:case2}. The results reveal that Qwen2.5-3B-Instruct exhibits clear deficiencies in medical mathematical reasoning. Specifically, its formula recall is unstable, often leading to omissions or incorrect invocation of standard clinical equations. In addition, the generated reasoning chains are relatively short and lack step-by-step calculations, which undermines transparency and auditability. Such limitations are particularly concerning in high-risk clinical tasks, where the absence of explicit intermediate steps makes errors more difficult to identify and correct.

In contrast, our proposed \textsc{MedCalc-R1}, after incorporating the knowledge-guided reward framework, demonstrates substantial advantages. (1) In terms of formula recall, the model reliably generates clinically valid formulas while avoiding common omissions and misapplications. (2) Regarding the reasoning process, \textsc{MedCalc-R1} produces more complete chains, encompassing explicit formula substitution, progressive calculations, and the final result, thereby enhancing transparency and interpretability. (3) For numerical accuracy, the model not only produces correct final answers but also ensures consistency across intermediate steps, significantly reducing potential computational errors.

\begin{table*}[]
    \centering
    \begin{tabular}{p{15.5cm}}
        \midrule
        You are a helpful assistant. The user asks a medical calculation question, and the Assistant solves it. The assistant first recalls the required formulas or scoring standards and thinks about the reasoning process in the mind, and then provides the user with the answer. The formulas, reasoning process, and final answer are enclosed within \formula{}, \think{} and \answer{} tags, respectively, i.e.,  \formula{medical formulas or scoring standards}  \think{reasoning process here} \answer{answer here}. Now, the user will provide you with key information about a patient and ask you to solve a calculation reasoning problem based on that information. After thinking, when you finally arrive at an answer, place the result within \answer{} tags. Here is the patient key note: \textcolor{red}{\{patient\_key\_note\}}. Here is the question: \textcolor{red}{\{question\}}. Let me solve this step by step.\\
        \midrule
    \end{tabular}
    \vspace{-2mm}
    \caption{Template for \textcolor{red}{question} and \textcolor{red}{patient\_key\_note} will be replaced with the specific question and patient note during training and inference.}
    \label{tab:instruction}
\end{table*}

\begin{algorithm*}[h]
\caption{Knowledge-Guided Training with GRPO}
\label{alg:training}
\begin{algorithmic}[1]
\Require Dataset $\mathcal{D}=\{(x, v^*, [L,U])\}$ with input $x=(c,q)$, ground-truth value $v^*$ and clinical safety interval $[L,U]$; policy $\pi_\theta$; reference policy $\pi_{\text{ref}}$; weights $\alpha,\beta,\gamma$; temperature $\tau$; group size $K$; GRPO clip $\epsilon$; KL weight $\lambda_{\text{KL}}$
\Ensure Trained policy $\pi_{\theta'}$
\Statex
\State \textbf{Stage I: Supervised Fine-Tuning (SFT)}
\For{epoch $=1,\ldots,E_{\text{SFT}}$}
    \For{mini-batch $\mathcal{B}\subset\mathcal{D}$}
        \State Update $\theta \leftarrow \theta - \eta \nabla_\theta \big[-\!\!\!\sum_{(x,y^*)\in\mathcal{B}}\sum_t \log \pi_\theta(y_t^* \mid y^*_{<t}, x)\big]$
    \EndFor
\EndFor
\Statex
\State \textbf{Stage II: RL with Knowledge-Guided Reward (GRPO)}
\For{epoch $=1,\ldots,E_{\text{RL}}$}
    \For{mini-batch $\mathcal{B}\subset\mathcal{D}$}
        \For{each $x\in\mathcal{B}$}
            \State Sample a group of $K$ candidates $\{y_i=(f_i,v_i)\}_{i=1}^K \sim \pi_\theta(\cdot\mid x)$
            \For{$i=1$ to $K$} \label{line:reward_begin}
                \State $R_{\text{format}} \gets \textsc{FormatReward}(y_i)$
                \State $R_{\text{knowledge}} \gets \textsc{KnowledgeReward}(f_i, x)$
                \State $R_{\text{answer}} \gets \textsc{AnswerReward}(v_i, v^*, [L,U], \tau)$
                \State $R_i \gets \alpha R_{\text{format}} + \beta R_{\text{knowledge}} + \gamma R_{\text{answer}}$ \label{line:reward_end}
            \EndFor
            \State $\bar{R} \gets \frac{1}{K}\sum_{i=1}^K R_i$ \Comment{group baseline}
            \For{$i=1$ to $K$}
                \State $A_i \gets R_i - \bar{R}$ \Comment{group-relative advantage}
                \State $r_i \gets \frac{\pi_\theta(y_i\mid x)}{\pi_{\theta_{\text{old}}}(y_i\mid x)}$ \Comment{importance ratio}
                \State $r_i^{\text{clip}} \gets \text{clip}(r_i, 1-\epsilon, 1+\epsilon)$
                \State $\mathcal{L}_i \gets -\min\{ r_i A_i,\; r_i^{\text{clip}} A_i \} + \lambda_{\text{KL}}\, \mathrm{KL}\!\big(\pi_\theta(\cdot\mid x)\,\|\,\pi_{\text{ref}}(\cdot\mid x)\big)$
            \EndFor
            \State Update $\theta \leftarrow \theta - \eta \nabla_\theta \frac{1}{K}\sum_{i=1}^K \mathcal{L}_i$
        \EndFor
    \EndFor
\EndFor
\State \Return $\pi_{\theta'}$
\end{algorithmic}
\end{algorithm*}

%% file: custom.bib
@inproceedings{tan2025reasonrft,
title={Reason-{RFT}: Reinforcement Fine-Tuning for Visual Reasoning of Vision Language Models},
author={Huajie Tan and Yuheng Ji and Xiaoshuai Hao and Xiansheng Chen and Pengwei Wang and Zhongyuan Wang and Shanghang Zhang},
booktitle={The Thirty-ninth Annual Conference on Neural Information Processing Systems},
year={2025},
url={https://openreview.net/forum?id=NdScoAix25}
}

@inproceedings{liu-etal-2025-compassverifier,
    title = "{C}ompass{V}erifier: A Unified and Robust Verifier for {LLM}s Evaluation and Outcome Reward",
    author = "Liu, Shudong  and
      Liu, Hongwei  and
      Liu, Junnan  and
      Xiao, Linchen  and
      Gao, Songyang  and
      Lyu, Chengqi  and
      Gu, Yuzhe  and
      Zhang, Wenwei  and
      Wong, Derek F.  and
      Zhang, Songyang  and
      Chen, Kai",
    booktitle = "Proceedings of the 2025 Conference on Empirical Methods in Natural Language Processing",
    month = nov,
    year = "2025",
    address = "Suzhou, China",
    publisher = "Association for Computational Linguistics",
    url = "https://aclanthology.org/2025.emnlp-main.1698/",
    doi = "10.18653/v1/2025.emnlp-main.1698",
    pages = "33454--33482"
}

@inproceedings{ahmadian-etal-2024-back,
    title = "Back to Basics: Revisiting {REINFORCE}-Style Optimization for Learning from Human Feedback in {LLM}s",
    author = {Ahmadian, Arash  and
      Cremer, Chris  and
      Gall{\'e}, Matthias  and
      Fadaee, Marzieh  and
      Kreutzer, Julia  and
      Pietquin, Olivier  and
      {\"U}st{\"u}n, Ahmet  and
      Hooker, Sara},
    booktitle = "Proceedings of the 62nd Annual Meeting of the Association for Computational Linguistics (Volume 1: Long Papers)",
    month = aug,
    year = "2024",
    address = "Bangkok, Thailand",
    publisher = "Association for Computational Linguistics",
    url = "https://aclanthology.org/2024.acl-long.662/",
    doi = "10.18653/v1/2024.acl-long.662",
    pages = "12248--12267"
}

@inproceedings{havrilla2024glore,
  title={GLoRe: when, where, and how to improve LLM reasoning via global and local refinements},
  author={Havrilla, Alex and Raparthy, Sharath and Nalmpantis, Christoforos and Dwivedi-Yu, Jane and Zhuravynski, Maksym and Hambro, Eric and Raileanu, Roberta},
  booktitle={Proceedings of the 41st International Conference on Machine Learning},
  pages={17719--17733},
  year={2024}
}

@article{lai2025med,
  title={Med-r1: Reinforcement learning for generalizable medical reasoning in vision-language models},
  author={Lai, Yuxiang and Zhong, Jike and Li, Ming and Zhao, Shitian and Li, Yuheng and Psounis, Konstantinos and Yang, Xiaofeng},
  journal={arXiv preprint arXiv:2503.13939},
  year={2025}
}

@article{su2025crossing,
  title={Crossing the Reward Bridge: Expanding RL with Verifiable Rewards Across Diverse Domains},
  author={Su, Yi and Yu, Dian and Song, Linfeng and Li, Juntao and Mi, Haitao and Tu, Zhaopeng and Zhang, Min and Yu, Dong},
  journal={arXiv preprint arXiv:2503.23829},
  year={2025}
}

@article{hurst2024gpt4o,
  title={Gpt-4o system card},
  author={Hurst, Aaron and Lerer, Adam and Goucher, Adam P and Perelman, Adam and Ramesh, Aditya and Clark, Aidan and Ostrow, AJ and Welihinda, Akila and Hayes, Alan and Radford, Alec and others},
  journal={arXiv preprint arXiv:2410.21276},
  year={2024}
}

@inproceedings{zhu2025menti,
  title={MeNTi: Bridging medical calculator and LLM agent with nested tool calling},
  author={Zhu, Yakun and Wei, Shaohang and Wang, Xu and Xue, Kui and Zhang, Shaoting and Zhang, Xiaofan},
  booktitle={Proceedings of the 2025 Conference of the Nations of the Americas Chapter of the Association for Computational Linguistics: Human Language Technologies (Volume 1: Long Papers)},
  pages={5097--5116},
  year={2025}
}

@inproceedings{rein2024gpqa,
title={{GPQA}: A Graduate-Level Google-Proof Q\&A Benchmark},
author={David Rein and Betty Li Hou and Asa Cooper Stickland and Jackson Petty and Richard Yuanzhe Pang and Julien Dirani and Julian Michael and Samuel R. Bowman},
booktitle={First Conference on Language Modeling},
year={2024},
url={https://openreview.net/forum?id=Ti67584b98}
}

@article{guo2025deepseek,
  title={Deepseek-r1: Incentivizing reasoning capability in llms via reinforcement learning},
  author={Guo, Daya and Yang, Dejian and Zhang, Haowei and Song, Junxiao and Zhang, Ruoyu and Xu, Runxin and Zhu, Qihao and Ma, Shirong and Wang, Peiyi and Bi, Xiao and others},
  journal={arXiv preprint arXiv:2501.12948},
  year={2025}
}

@article{chen2024huatuogpto1,
  title={Huatuogpt-o1, towards medical complex reasoning with llms},
  author={Chen, Junying and Cai, Zhenyang and Ji, Ke and Wang, Xidong and Liu, Wanlong and Wang, Rongsheng and Hou, Jianye and Wang, Benyou},
  journal={arXiv preprint arXiv:2412.18925},
  year={2024}
}

@inproceedings{chu-etal-2024-navigate,
    title = "Navigate through Enigmatic Labyrinth A Survey of Chain of Thought Reasoning: Advances, Frontiers and Future",
    author = "Chu, Zheng  and
      Chen, Jingchang  and
      Chen, Qianglong  and
      Yu, Weijiang  and
      He, Tao  and
      Wang, Haotian  and
      Peng, Weihua  and
      Liu, Ming  and
      Qin, Bing  and
      Liu, Ting",
    booktitle = "Proceedings of the 62nd Annual Meeting of the Association for Computational Linguistics (Volume 1: Long Papers)",
    month = aug,
    year = "2024",
    pages = "1173--1203"
}

@inproceedings{khandekar2024medcalcbench,
title={MedCalc-Bench: Evaluating Large Language Models for Medical Calculations},
author={Nikhil Khandekar and Qiao Jin and Guangzhi Xiong and Soren Dunn and Serina S Applebaum and Zain Anwar and Maame Sarfo-Gyamfi and Conrad W Safranek and Abid Anwar and Andrew Jiaxing Zhang and Aidan Gilson and Maxwell B Singer and Amisha D Dave and R. Andrew Taylor and Aidong Zhang and Qingyu Chen and Zhiyong Lu},
booktitle={The Thirty-eight Conference on Neural Information Processing Systems Datasets and Benchmarks Track},
year={2024},
url={https://openreview.net/forum?id=VXohja0vrQ}
}

@article{shao2024deepseekmath,
  title={Deepseekmath: Pushing the limits of mathematical reasoning in open language models},
  author={Shao, Zhihong and Wang, Peiyi and Zhu, Qihao and Xu, Runxin and Song, Junxiao and Bi, Xiao and Zhang, Haowei and Zhang, Mingchuan and Li, YK and others},
  journal={arXiv preprint arXiv:2402.03300},
  year={2024}
}

@inproceedings{yang2024harnessing,
  title={Harnessing the Power of Large Language Models for Natural Language to First-Order Logic Translation},
  author={Yang, Yu’an and Xiong, Siheng and Payani, Ali and Shareghi, Ehsan and Fekri, Faramarz},
  booktitle={Proceedings of the 62nd Annual Meeting of the Association for Computational Linguistics (Volume 1: Long Papers)},
  pages={6942--6959},
  year={2024}
}

@article{wang2024reinforcement,
  title={Reinforcement learning enhanced llms: A survey},
  author={Wang, Shuhe and Zhang, Shengyu and Zhang, Jie and Hu, Runyi and Li, Xiaoya and Zhang, Tianwei and Li, Jiwei and Wu, Fei and Wang, Guoyin and Hovy, Eduard},
  journal={arXiv preprint arXiv:2412.10400},
  year={2024}
}

@article{qin2024relevant,
  title={Relevant or Random: Can LLMs Truly Perform Analogical Reasoning?},
  author={Qin, Chengwei and Xia, Wenhan and Wang, Tan and Jiao, Fangkai and Hu, Yuchen and Ding, Bosheng and Chen, Ruirui and Joty, Shafiq},
  journal={arXiv preprint arXiv:2404.12728},
  year={2024}
}

@article{guo2024deepseek,
  title={DeepSeek-Coder: When the Large Language Model Meets Programming--The Rise of Code Intelligence},
  author={Guo, Daya and Zhu, Qihao and Yang, Dejian and Xie, Zhenda and Dong, Kai and Zhang, Wentao and Chen, Guanting and Bi, Xiao and Wu, Yu and Li, YK and others},
  journal={arXiv preprint arXiv:2401.14196},
  year={2024}
}

@article{austin2021program,
  title={Program synthesis with large language models},
  author={Austin, Jacob and Odena, Augustus and Nye, Maxwell and Bosma, Maarten and Michalewski, Henryk and Dohan, David and Jiang, Ellen and Cai, Carrie and Terry, Michael and Le, Quoc and others},
  journal={arXiv preprint arXiv:2108.07732},
  year={2021}
}

@article{liu2025logical,
  title={Logical reasoning in large language models: A survey},
  author={Liu, Hanmeng and Fu, Zhizhang and Ding, Mengru and Ning, Ruoxi and Zhang, Chaoli and Liu, Xiaozhang and Zhang, Yue},
  journal={arXiv preprint arXiv:2502.09100},
  year={2025}
}

@article{yang2024qwen2,
  title={Qwen2. 5-math technical report: Toward mathematical expert model via self-improvement},
  author={Yang, An and Zhang, Beichen and Hui, Binyuan and Gao, Bofei and Yu, Bowen and Li, Chengpeng and Liu, Dayiheng and Tu, Jianhong and Zhou, Jingren and Lin, Junyang and others},
  journal={arXiv preprint arXiv:2409.12122},
  year={2024}
}

@inproceedings{wei2022chain,
title={Chain of Thought Prompting Elicits Reasoning in Large Language Models},
author={Jason Wei and Xuezhi Wang and Dale Schuurmans and Maarten Bosma and brian ichter and Fei Xia and Ed H. Chi and Quoc V Le and Denny Zhou},
booktitle={Advances in Neural Information Processing Systems},
editor={Alice H. Oh and Alekh Agarwal and Danielle Belgrave and Kyunghyun Cho},
year={2022},
url={https://openreview.net/forum?id=_VjQlMeSB_J}
}

@article{qu2025tool,
  title={Tool learning with large language models: A survey},
  author={Qu, Changle and Dai, Sunhao and Wei, Xiaochi and Cai, Hengyi and Wang, Shuaiqiang and Yin, Dawei and Xu, Jun and Wen, Ji-Rong},
  journal={Frontiers of Computer Science},
  volume={19},
  number={8},
  pages={198343},
  year={2025},
  publisher={Springer}
}

@article{wang2025survey,
  title={A Survey on Large Language Models for Mathematical Reasoning},
  author={Wang, Peng-Yuan and Liu, Tian-Shuo and Wang, Chenyang and Wang, Yi-Di and Yan, Shu and Jia, Cheng-Xing and Liu, Xu-Hui and Chen, Xin-Wei and Xu, Jia-Cheng and Li, Ziniu and others},
  journal={arXiv preprint arXiv:2506.08446},
  year={2025}
}

@inproceedings{zhang2025generative,
title={Generative Verifiers: Reward Modeling as Next-Token Prediction},
author={Lunjun Zhang and Arian Hosseini and Hritik Bansal and Mehran Kazemi and Aviral Kumar and Rishabh Agarwal},
booktitle={The Thirteenth International Conference on Learning Representations},
year={2025},
url={https://openreview.net/forum?id=Ccwp4tFEtE}
}

@article{cobbe2021training,
  title={Training verifiers to solve math word problems},
  author={Cobbe, Karl and Kosaraju, Vineet and Bavarian, Mohammad and Chen, Mark and Jun, Heewoo and Kaiser, Lukasz and Plappert, Matthias and Tworek, Jerry and Hilton, Jacob and Nakano, Reiichiro and others},
  journal={arXiv preprint arXiv:2110.14168},
  year={2021}
}

@inproceedings{hendrycks2021measuring,
title={Measuring Mathematical Problem Solving With the {MATH} Dataset},
author={Dan Hendrycks and Collin Burns and Saurav Kadavath and Akul Arora and Steven Basart and Eric Tang and Dawn Song and Jacob Steinhardt},
booktitle={Thirty-fifth Conference on Neural Information Processing Systems Datasets and Benchmarks Track (Round 2)},
year={2021},
url={https://openreview.net/forum?id=7Bywt2mQsCe}
}

@inproceedings{gao2023pal,
  title={Pal: Program-aided language models},
  author={Gao, Luyu and Madaan, Aman and Zhou, Shuyan and Alon, Uri and Liu, Pengfei and Yang, Yiming and Callan, Jamie and Neubig, Graham},
  booktitle={International Conference on Machine Learning},
  pages={10764--10799},
  year={2023},
  organization={PMLR}
}

@inproceedings{imani2023mathprompter,
  title={MathPrompter: Mathematical Reasoning using Large Language Models},
  author={Imani, Shima and Du, Liang and Shrivastava, Harsh},
  booktitle={Proceedings of the 61st Annual Meeting of the Association for Computational Linguistics (Volume 5: Industry Track)},
  pages={37--42},
  year={2023}
}

@article{ouyang2022training,
  title={Training language models to follow instructions with human feedback},
  author={Ouyang, Long and Wu, Jeffrey and Jiang, Xu and Almeida, Diogo and Wainwright, Carroll and Mishkin, Pamela and Zhang, Chong and Agarwal, Sandhini and Slama, Katarina and Ray, Alex and others},
  journal={Advances in neural information processing systems},
  volume={35},
  pages={27730--27744},
  year={2022}
}

@inproceedings{lightman2024lets,
title={Let's Verify Step by Step},
author={Hunter Lightman and Vineet Kosaraju and Yuri Burda and Harrison Edwards and Bowen Baker and Teddy Lee and Jan Leike and John Schulman and Ilya Sutskever and Karl Cobbe},
booktitle={The Twelfth International Conference on Learning Representations},
year={2024},
url={https://openreview.net/forum?id=v8L0pN6EOi}
}

@article{yuan2023rrhf,
  title={Rrhf: Rank responses to align language models with human feedback},
  author={Yuan, Hongyi and Yuan, Zheng and Tan, Chuanqi and Wang, Wei and Huang, Songfang and Huang, Fei},
  journal={Advances in Neural Information Processing Systems},
  volume={36},
  pages={10935--10950},
  year={2023}
}

@article{singhal2023large,
  title={Large language models encode clinical knowledge},
  author={Singhal, Karan and Azizi, Shekoofeh and Tu, Tao and Mahdavi, S Sara and Wei, Jason and Chung, Hyung Won and Scales, Nathan and Tanwani, Ajay and Cole-Lewis, Heather and Pfohl, Stephen and others},
  journal={Nature},
  volume={620},
  number={7972},
  pages={172--180},
  year={2023},
  publisher={Nature Publishing Group}
}

@article{lucas2024reasoning,
  title={Reasoning with large language models for medical question answering},
  author={Lucas, Mary M and Yang, Justin and Pomeroy, Jon K and Yang, Christopher C},
  journal={Journal of the American Medical Informatics Association},
  volume={31},
  number={9},
  pages={1964--1975},
  year={2024},
  publisher={Oxford University Press}
}

@misc{hu2025reinforce,
      title={REINFORCE++: Stabilizing Critic-Free Policy Optimization with Global Advantage Normalization}, 
      author={Jian Hu and Jason Klein Liu and Haotian Xu and Wei Shen},
      year={2025},
      eprint={2501.03262},
      archivePrefix={arXiv},
      primaryClass={cs.CL},
      url={https://arxiv.org/abs/2501.03262}, 
}

@article{schulman2017proximal,
  title={Proximal policy optimization algorithms},
  author={Schulman, John and Wolski, Filip and Dhariwal, Prafulla and Radford, Alec and Klimov, Oleg},
  journal={arXiv preprint arXiv:1707.06347},
  year={2017}
}

@article{jaech2024openai,
  title={Openai o1 system card},
  author={Jaech, Aaron and Kalai, Adam and Lerer, Adam and Richardson, Adam and El-Kishky, Ahmed and Low, Aiden and Helyar, Alec and Madry, Aleksander and Beutel, Alex and Carney, Alex and others},
  journal={arXiv preprint arXiv:2412.16720},
  year={2024}
}

@article{thirunavukarasu2023large,
  title={Large language models in medicine},
  author={Thirunavukarasu, Arun James and Ting, Darren Shu Jeng and Elangovan, Kabilan and Gutierrez, Laura and Tan, Ting Fang and Ting, Daniel Shu Wei},
  journal={Nature medicine},
  volume={29},
  number={8},
  pages={1930--1940},
  year={2023},
  publisher={Nature Publishing Group US New York}
}

@article{cicero2020medication,
  title={Medication dosing safety for pediatric patients: recognizing gaps, safety threats, and best practices in the emergency medical services setting. A position statement and resource document from NAEMSP},
  author={Cicero, Mark X and Adelgais, Kathleen and Hoyle, John D and Lyng, John W and Harris, Matthew and Moore, Brian and Gausche-Hill, Marianne and Pediatric Committee of NAEMSP Adopted by NAEMSP Board of Directors},
  journal={Prehospital Emergency Care},
  volume={25},
  number={2},
  pages={294--306},
  year={2020},
  publisher={Taylor \& Francis}
}

@article{hijji2025indispensable,
  title={An Indispensable Requirement for Medical Dosage Calculation: Basic Mathematical Skills of Baccalaureate Nursing Students},
  author={Hijji, Belal Mahmoud},
  journal={Nursing Reports},
  volume={15},
  number={5},
  pages={150},
  year={2025},
  publisher={MDPI}
}

@article{xu2025towards,
  title={Towards large reasoning models: A survey of reinforced reasoning with large language models},
  author={Xu, Fengli and Hao, Qianyue and Zong, Zefang and Wang, Jingwei and Zhang, Yunke and Wang, Jingyi and Lan, Xiaochong and Gong, Jiahui and Ouyang, Tianjian and Meng, Fanjin and others},
  journal={arXiv preprint arXiv:2501.09686},
  year={2025}
}

@inproceedings{kim-yoon-2025-questioning,
    title = "Questioning Our Questions: How Well Do Medical {QA} Benchmarks Evaluate Clinical Capabilities of Language Models?",
    author = "Kim, Siun  and
      Yoon, Hyung-Jin",
    booktitle = "Proceedings of the 24th Workshop on Biomedical Language Processing",
    month = aug,
    year = "2025",
    address = "Viena, Austria",
    publisher = "Association for Computational Linguistics",
    url = "https://aclanthology.org/2025.bionlp-1.24/",
    doi = "10.18653/v1/2025.bionlp-1.24",
    pages = "274--296",
    ISBN = "979-8-89176-275-6"
}

@inproceedings{ruano-etal-2025-effective,
    title = "Effective Multi-Task Learning for Biomedical Named Entity Recognition",
    author = "Ruano, Jo{\~a}o  and
      Correia, Gon{\c{c}}alo  and
      Barreiros, Leonor  and
      Mendes, Afonso",
    booktitle = "Proceedings of the 24th Workshop on Biomedical Language Processing",
    month = aug,
    year = "2025",
    address = "Viena, Austria",
    publisher = "Association for Computational Linguistics",
    url = "https://aclanthology.org/2025.bionlp-1.20/",
    doi = "10.18653/v1/2025.bionlp-1.20",
    pages = "225--239",
    ISBN = "979-8-89176-275-6"
}

@inproceedings{bahrololloomi-etal-2025-transformer,
    title = "Transformer-Based Medical Statement Classification in Doctor-Patient Dialogues",
    author = "Bahrololloomi, Farnod  and
      Luderschmidt, Johannes  and
      Fu, Biying",
    booktitle = "Proceedings of the 24th Workshop on Biomedical Language Processing",
    month = aug,
    year = "2025",
    address = "Viena, Austria",
    publisher = "Association for Computational Linguistics",
    url = "https://aclanthology.org/2025.bionlp-1.7/",
    doi = "10.18653/v1/2025.bionlp-1.7",
    pages = "63--73",
    ISBN = "979-8-89176-275-6"
}

@inproceedings{he2025breaking,
  title={Breaking the Reasoning Barrier A Survey on LLM Complex Reasoning through the Lens of Self-Evolution},
  author={He, Tao and Li, Hao and Chen, Jingchang and Liu, Runxuan and Cao, Yixin and Liao, Lizi and Zheng, Zihao and Chu, Zheng and Liang, Jiafeng and Liu, Ming and others},
  booktitle={Findings of the Association for Computational Linguistics: ACL 2025},
  pages={7377--7417},
  year={2025}
}

@article{qwen2.5,
    title   = {Qwen2.5 Technical Report}, 
    author  = {An Yang and Baosong Yang and Beichen Zhang and Binyuan Hui and Bo Zheng and Bowen Yu and Chengyuan Li and Dayiheng Liu and Fei Huang and Haoran Wei and Huan Lin and Jian Yang and Jianhong Tu and Jianwei Zhang and Jianxin Yang and Jiaxi Yang and Jingren Zhou and Junyang Lin and Kai Dang and Keming Lu and Keqin Bao and Kexin Yang and Le Yu and Mei Li and Mingfeng Xue and Pei Zhang and Qin Zhu and Rui Men and Runji Lin and Tianhao Li and Tingyu Xia and Xingzhang Ren and Xuancheng Ren and Yang Fan and Yang Su and Yichang Zhang and Yu Wan and Yuqiong Liu and Zeyu Cui and Zhenru Zhang and Zihan Qiu},
    journal = {arXiv preprint arXiv:2412.15115},
    year    = {2024}
}

@misc{qwq32b,
    title = {QwQ-32B: Embracing the Power of Reinforcement Learning},
    url = {https://qwenlm.github.io/blog/qwq-32b/},
    author = {Qwen Team},
    month = {March},
    year = {2025}
}
